\documentclass{article}

\usepackage[final]{neurips_2019}

\usepackage[utf8]{inputenc} 
\usepackage[T1]{fontenc}    
\usepackage{hyperref}       
\usepackage{url}            
\usepackage{booktabs}       
\usepackage{amsfonts}       
\usepackage{nicefrac}       
\usepackage{microtype}      
\usepackage{floatrow}
\usepackage[normalem]{ulem}
\setcitestyle{authoryear}
\usepackage{wrapfig}
\usepackage{placeins}

\usepackage{algorithm}
\usepackage{algorithmic}

\usepackage{xcolor}
\usepackage{amsmath}
\usepackage{amssymb}
\usepackage{chngcntr}
\usepackage{amsthm}
\usepackage{ifthen}
\usepackage{graphicx}
\usepackage{caption}
\usepackage{subcaption}
\usepackage{mathtools}

\makeatletter
\newtheorem*{rep@theorem}{\rep@title}
\newcommand{\newreptheorem}[2]{%
\newenvironment{rep#1}[1]{%
 \def\rep@title{#2 \ref{##1}}%
 \begin{rep@theorem}}%
 {\end{rep@theorem}}}
\makeatother

\newtheorem{lemma}{Lemma}
\newtheorem{proposition}{Proposition}
\newtheorem{theorem}{Theorem}

\newtheorem{corollary}{Corollary}
\newtheorem{assumption}{Assumption}
\newreptheorem{lemma}{Lemma}
\newreptheorem{proposition}{Proposition}
\newreptheorem{theorem}{Theorem}
\newreptheorem{defn}{Definition}
\newreptheorem{conjecture}{Conjecture}
\newreptheorem{corollary}{Corollary}
\newreptheorem{assumption}{Assumption}

\newcommand\numberthis{\addtocounter{equation}{1}\tag{\theequation}}

\newcommand{\hide}[1]{}

\newcommand\sI{\ensuremath{\mathcal{I}}}

\newcommand\sX{\ensuremath{\mathcal{X}}}
\newcommand\sY{\ensuremath{\mathcal{Y}}}

\DeclareMathOperator*{\tr}{tr}

\newcommand\inv[1]{\ensuremath{\frac{1}{#1}}}

\newcommand\R{\ensuremath{\mathbb{R}}} 
\newcommand\eqdef{\ensuremath{\stackrel{\rm def}{=}}} 
\newcommand{\1}{\mathbb{I}} 
\newcommand{\bone}{\mathbf{1}} 
\newcommand\refeqn[1]{(\ref{eqn:#1})}

\newcommand\refsec[1]{Section~\ref{sec:#1}}

\newcommand\reffig[1]{Figure~\ref{fig:#1}}

\newcommand\reftab[1]{Table~\ref{tab:#1}}
\newcommand\refapp[1]{Appendix~\ref{sec:#1}}

\newcommand\reflem[1]{Lemma~\ref{lem:#1}}

\newcommand\refprop[1]{Proposition~\ref{prop:#1}}

\newcommand\refcor[1]{Corollary~\ref{cor:#1}}
\newcommand\refassumps[2]{Assumptions~\ref{assump:#1} and ~\ref{assump:#2}}

\ifthenelse{\isundefined{\definition}}{\newtheorem{definition}{Definition}}{}
\ifthenelse{\isundefined{\assumption}}{\newtheorem{assumption}{Assumption}}{}
\ifthenelse{\isundefined{\hypothesis}}{\newtheorem{hypothesis}{Hypothesis}}{}
\ifthenelse{\isundefined{\proposition}}{\newtheorem{proposition}{Proposition}}{}
\ifthenelse{\isundefined{\theorem}}{\newtheorem{theorem}{Theorem}}{}
\ifthenelse{\isundefined{\lemma}}{\newtheorem{lemma}{Lemma}}{}
\ifthenelse{\isundefined{\corollary}}{\newtheorem{corollary}{Corollary}}{}
\ifthenelse{\isundefined{\alg}}{\newtheorem{alg}{Algorithm}}{}
\ifthenelse{\isundefined{\example}}{\newtheorem{example}{Example}}{}

\newcommand{\nt}{\nabla_\theta}

\newcommand{\Ho}{H_{\lambda, \bone}}

\newcommand{\lmin}{\sigma_{\mathrm{min}}}
\newcommand{\lmax}{\sigma_{\mathrm{max}}}
\newcommand{\opnorm}[1]{\left\|#1\right\|_\mathrm{op}}
\newcommand{\twonorm}[1]{\left\|#1\right\|_2}
\newcommand{\onenorm}[1]{\left\|#1\right\|_1}

\newcommand{\Cmax}{C_\mathrm{max}}

\newcommand{\nI}{\sI^\mathrm{Nt}_f}
\newcommand{\actI}{\sI^{*}_f}
\newcommand{\predI}{\sI_f}
\newcommand{\Errn}{\mathrm{Err}_\mathrm{Nt\text{-}act}}
\newcommand{\Errni}{\mathrm{Err}_\mathrm{Nt\text{-}inf}}
\newcommand{\htn}{\hat\theta_{\mathrm{Nt}}}
\newcommand{\dnw}{\Delta\theta_{\mathrm{Nt}}(w)}
\newcommand{\diw}{\Delta\theta_{\mathrm{inf}}(w)}

\newcommand{\Errf}{\mathrm{Err}_{f,3}}
\newcommand{\tErrf}{\mathrm{Err}_{f,2}}

\newcommand{\vtest}{{v_{\mathrm{test}}}}

\newcommand{\xtest}{{x_{\mathrm{test}}}}
\newcommand{\ytest}{{y_{\mathrm{test}}}}

\newcommand{\ow}{{\bone - w}}

\newcommand{\hto}{\hat\theta(\bone)}
\newcommand{\htow}{\hat\theta(\bone - w)}
\newcommand{\htotw}{\hat\theta(\bone - tw)}

\newcommand{\abs}[1]{\left\lvert#1\right\rvert}

\newcommand\todo[1]{}
\newcommand\pw[1]{}
\newcommand\ka[1]{}
\newcommand\hu[1]{}
\newcommand\pl[1]{}
\newcommand\pldel[1]{}

\author{
  Pang Wei Koh\thanks{Equal contribution.}
  \qquad Kai-Siang Ang\footnotemark[1]
  \qquad Hubert H.~K.~Teo\footnotemark[1]
  \qquad Percy Liang\\
  Department of Computer Science\\
  Stanford University\\
  \texttt{\{pangwei@cs, kaiang@, hteo@, pliang@cs\}.stanford.edu}
}

\begin{document}

\title{On the Accuracy of Influence Functions \\ for Measuring Group Effects}

\maketitle
\begin{abstract}
Influence functions estimate the effect of removing a training point on a model without the need to retrain.
They are based on a first-order Taylor approximation that is guaranteed to be accurate for sufficiently small changes to the model,
and so are commonly used to study the effect of individual points in large datasets.
However, we often want to study the effects of large \emph{groups} of training points, e.g.,
to diagnose batch effects or apportion credit between different data sources.
Removing such large groups can result in significant changes to the model.
Are influence functions still accurate in this setting?
In this paper, we find that across many different types of groups and for a range of real-world datasets,
the predicted effect (using influence functions) of a group correlates surprisingly well with its actual effect,
even if the absolute and relative errors are large.
Our theoretical analysis shows that such strong correlation arises only under certain settings and
need not hold in general,
indicating that real-world datasets have particular properties that
allow the influence approximation to be accurate.

\end{abstract}

\section{Introduction}\label{sec:intro}

Influence functions \citep{jaeckel1972infinitesimal, hampel1974influence, cook1977detection} estimate the effect of removing an
individual training point on a model's predictions without the computationally-prohibitive cost of retraining the model.
Tracing a model's output back to its training data can be useful:
influence functions have been recently applied to explain predictions \citep{koh2017understanding},
produce confidence intervals \citep{schulam2019can},
investigate model bias \citep{brunet2018understanding, wang2019repairing},
improve human trust \citep{zhou2019effects},
and even craft data poisoning attacks \citep{koh2019stronger}.

Influence functions are based on first-order Taylor approximations that are accurate for estimating small perturbations to the model,
which makes them suitable for predicting the effects of removing individual training points on the model.
However, we often want to study the effects of removing \emph{groups} of points, which represent large perturbations to the data.
For example, we might wish to analyze the effect of data collected from different experimental batches \citep{leek2010tackling} or demographic groups \citep{chen2018my};
apportion credit between crowdworkers, each of whom generated part of the data \citep{arrieta2018should};
or, in a multi-party learning setting, ensure that no individual user has too much influence on the joint model \citep{hayes2018contamination}.
Are influence functions still accurate when predicting the effects of (removing) these larger groups?

In this paper, we first show empirically that on real datasets and across a broad variety of groups of data, the predicted and actual effects are strikingly \emph{correlated} (Spearman $\rho$ of $0.8$ to $1.0$),
such that the groups with the largest actual effect also tend to have the largest predicted effect. Moreover, the predicted effect tends to \emph{underestimate} the actual effect, suggesting that it could be an approximate lower bound in practice.
Using influence functions to predict the actual effect of removing large, coherent groups of data can therefore still be useful,
even though the violation of the small-perturbation assumption can result in high absolute and relative errors between the predicted and actual effects.

What explains these phenomena of correlation and underestimation?
Prior theoretical work focused on establishing the conditions under which this influence approximation is accurate, i.e., the error between the actual and predicted effects is small \citep{giordano2019swiss, rad2018scalable}.
However, in our setting of removing large, coherent groups of data,
this error can be quite large.
As a first step towards understanding the behavior of the influence approximation in this regime,
we characterize the relationship between the predicted and actual effects of a group via the one-step Newton approximation \citep{pregibon1981logistic},
which we find is a surprisingly accurate approximation in practice.
We show that correlation and underestimation arise under certain settings
(e.g., removing multiple copies of a single training point),
but need not hold in general,
which opens up the intriguing question of why we observe those phenomena across a wide range of empirical settings.

Finally, we exploit the correlation of predicted and actual group effects in two example case studies: a chemical-disease relationship (CDR) task, where the groups correspond to different labeling functions \citep{hancock2018babble}, and a natural language inference (NLI) task \citep{williams2018broad}, where the groups come from different crowdworkers.
On the CDR task, we find that the influence of each labeling function correlates with its size (the number of examples it labels) but not its average accuracy, which suggests that practitioners should focus on the coverage of the labeling functions they construct. In contrast, on the NLI task, we find that the influence of each crowdworker is uncorrelated with the number of examples they contibute, which suggests that practitioners should focus on how to elicit high-quality examples from crowdworkers over increasing quantity.

\section{Background and problem setup}\label{sec:setting}
Consider learning a predictive model with parameters $\theta \in \Theta$ that
maps from an input space $\sX$ to an output space $\sY$.
We are given $n$ training points $\{(x_1, y_1), \ldots, (x_n, y_n)\}$
and a loss function $\ell(x, y, \theta)$ that is twice-differentiable and convex in $\theta$.
To train the model, we select the model parameters
\begin{align}
  \hto = {\arg\min}_{\theta\in\Theta}  \left[ \sum_{i=1}^n \ell(x_i, y_i; \theta) \right] + \frac{\lambda}{2} \|\theta\|_2^2
\end{align}
that minimize the $L_2$-regularized empirical risk,
where $\lambda > 0$ controls regularization strength.
The all-ones vector $\bone$ in $\hto$ denotes that the initial training points all have uniform sample weights.

Our goal is to measure the effects of different groups of training data
on the model: if we removed a subset of training points $W$,
how much would the model $\hat\theta$ change?
Concretely, we define a vector $w \in \{0,1\}^n$ of sample weights with $w_i = \1((x_i, y_i) \in W)$
and consider the modified parameters
\begin{align}
\htow = {\arg\min}_{\theta\in\Theta}  \left[\sum_{i=1}^n (1 - w_i) \ell(x_i, y_i; \theta) \right] + \frac{\lambda}{2} \|\theta\|_2^2
\end{align}
corresponding to retraining the model after excluding $W$.
We refer to $w$ as the subset (corresponding to $W$);
the number of removed points as $\|w\|_1$;
and the fraction of removed points as $\alpha = \|w\|_1 / n$.

The \emph{actual effect} $\actI:[0,1]^n \to \R$ of the subset $w$ is
\begin{align}
  \actI(w) = f(\htow) - f(\hto),
\end{align}
where the evaluation function $f:\Theta\to\R$ measures a quantity of interest. Specifically, we study:
\begin{itemize}
  \item The \emph{change in test prediction}, with $f(\theta) = \theta^\top \xtest$. Linear models (for regression or binary classification) make predictions that are functions of $\theta^\top \xtest$, so this measures the effect that removing a subset will have on the model's prediction for some test point $\xtest$.

  \item The \emph{change in test loss}, with $f(\theta) = \ell(\xtest, \ytest; \theta)$, which is similar to the test prediction.

  \item The \emph{change in self-loss}, with $f(\theta) = \sum_{i=1}^n w_i \ell(x_i, y_i; \theta)$, measures the increase in loss on the removed points $w$. Its average over all subsets of size $\|w\|_1$ is the estimated extra loss that leave-$\|w\|_1$-out cross-validation (CV) measures over the training loss.

\end{itemize}

\subsection{Influence functions}
The issue with computing the actual effect $\actI(w)$ is that retraining the model to compute $\htow$ for each subset $w$ can be prohibitively expensive.
Influence functions provide a relatively efficient first-order approximation to $\actI(w)$ that avoids retraining.

Consider the function $q_w: [0, 1] \to \R$ with
$q_w(t) = f\bigl(\htotw\bigr)$,
such that the actual effect $\actI(w)$ can be written as $q_w(1) - q_w(0)$.
We define the \emph{predicted effect} of the subset $w$ to be
its \emph{influence} $\predI(w) = q_w'(0) \approx q_w(1) - q_w(0)$;
in this paper, we use the term predicted effect interchangeably with influence.
Intuitively, influence measures the effect of removing an infinitesimal weight from each point in $w$ and then linearly extrapolates to removing all of $w$.\footnote{In the statistics literature, influence typically refers to the effect of \emph{adding} weight, so the sign is flipped.} By taking a Taylor approximation (see, e.g., \citet{hampel1986robust} for details), the influence can be computed as
\begin{align}
\predI(w) \eqdef q_w'(0)\nonumber
&= \nt f\bigl(\hto\bigr)^\top \left[ \frac{d}{dt}\htotw\Bigr|_{\substack{t=0}} \right]\nonumber\\
&= \nt f\bigl(\hto\bigr)^\top \Ho^{-1} g_\bone(w),
\end{align}
where
$g_\bone(w) = \sum_{i=1}^n w_i\nt\ell(x_i, y_i; \hto)$,
$H_\bone = \sum_{i=1}^n \nt^2\ell(x_i, y_i; \hto)$,
and $\Ho = H_\bone + \lambda I$.
When measuring the change in test prediction or test loss, influence is additive: if $w = w_1 + w_2$, then $\predI(w) = \predI(w_1) + \predI(w_2)$, i.e., the influence of a subset is the sum of influences of its constituent points,
and we can efficiently compute the influence of any subset by pre-computing the influence of each individual point (e.g., by taking a single inverse Hessian-vector product, as in
\citet{koh2017understanding}).
However, when measuring the change in self-loss, influence is not additive and requires a separate calculation for each subset removed.

\subsection{Relation to prior work}\label{sec:related}
Influence functions---introduced in the seminal work of \citet{hampel1974influence} and in \citet{jaeckel1972infinitesimal}, where it was called the infinitesimal jackknife---have a rich history in robust statistics.
The use of influence functions in the ML community is more recent, though growing; in \refsec{intro}, we provide references for several recent applications of influence functions in ML.

Removing a single training point, especially when the total number of points $n$ is large, represents a small perturbation to the training distribution,
so we expect the first-order influence approximation to be accurate.
Indeed, prior work on the accuracy of influence has focused on this regime: e.g.,
\citet{debruyne2008model, liu2014efficient, rad2018scalable, giordano2019swiss} give
evidence that the influence on self-loss can approximate LOOCV, and \citet{koh2017understanding} similarly examined the accuracy of estimating the change in test loss after removing single training points.

However, removing a constant fraction $\alpha$ of the training data represents a large perturbation to the training distribution.
To the best of our knowledge,
this setting has
not been empirically studied;
perhaps the closest work is \citet{khanna2019interpreting}'s use of Bayesian quadrature to estimate a maximally influential subset.
Instead, older references have alluded to the phenomena of correlation and underestimation we observe:
\citet{pregibon1981logistic} note that influence tends to be conservative,
while \citet{hampel1986robust} say that ``bold extrapolations'' (i.e., large perturbations) are often still useful.
On the theoretical front, \citet{giordano2019swiss} established
finite-sample error bounds that apply to groups, e.g., showing that the leave-$k$-out approximation is consistent as the fraction of removed points $\alpha \to 0$.
Our focus is instead on the relationship of the actual effect $\actI(w)$ and predicted effect (influence) $\predI(w)$ in the regime where $\alpha$ is constant and
the error $|\actI(w) - \predI(w)|$ is large.

\section{Empirical accuracy of influence functions on constructed groups}\label{sec:accuracy}

How well do influence functions estimate the effect of (removing) a group of training points?
If $n$ is large and we remove a subset $w$ uniformly at random, the new parameters $\htow$ should remain close to $\hto$ even when if fraction of removed points $\alpha$ is non-negligible, so the influence error $|\actI(w) - \predI(w)|$ should be small.
However, we are usually interested in removing coherent, non-random groups, e.g., all points from a data source or share some feature. In such settings, the parameters $\htow$ and $\hto$ might differ substantially, and the error $|\actI(w) - \predI(w)|$ could be large.
Put another way, there could be a cluster of points such that removing one of those points would not change the model by much---so influence could be low---but removing all of them would.

\begin{figure}[t]
  \centering
  \includegraphics[width=\textwidth]{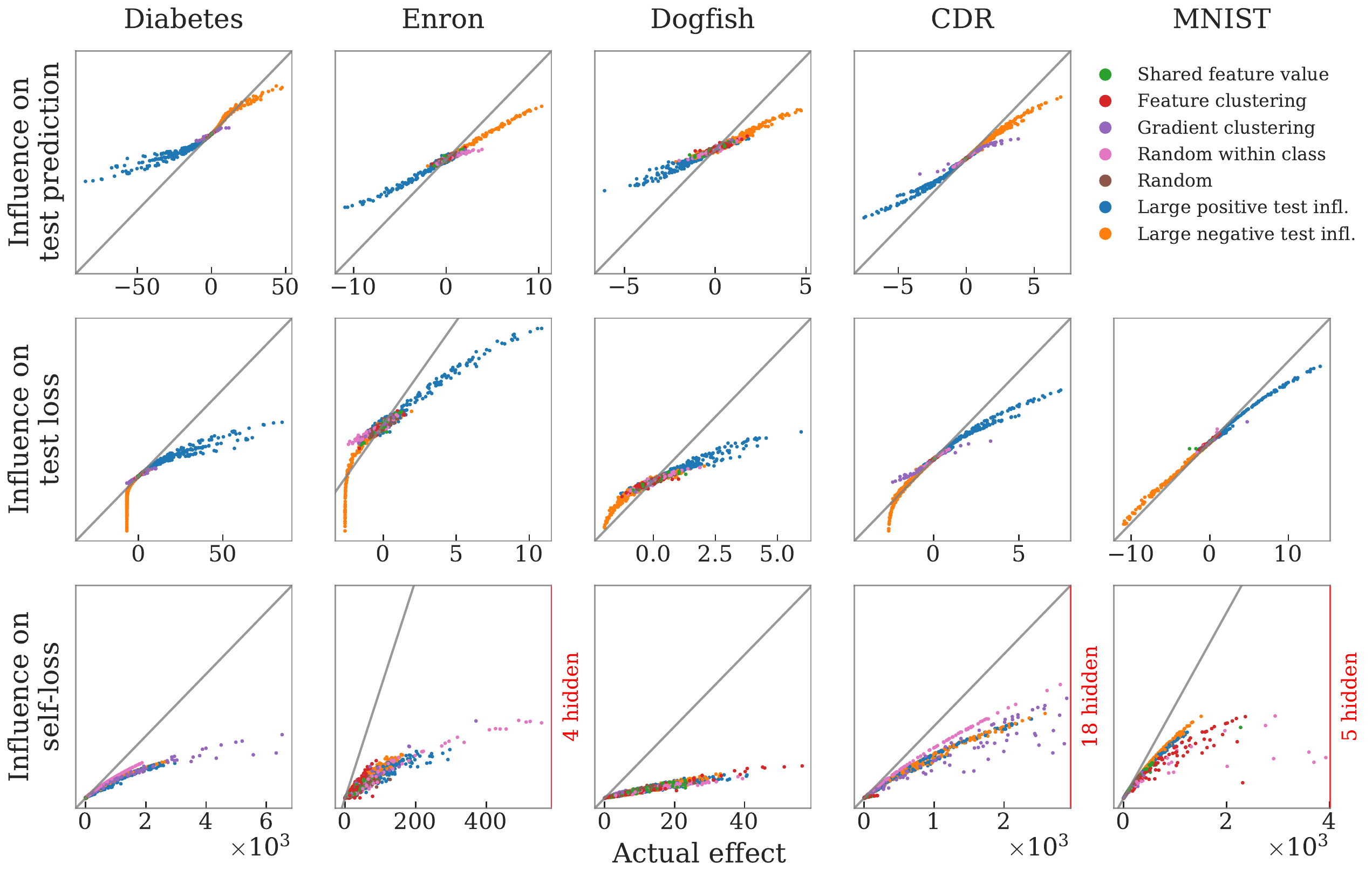}
  \vspace{3mm}
  \caption{
  Influences vs. actual effects of coherent groups of points ranging from $0.25\%$ to $25\%$ in size. Each point corresponds to a group, and its color reflects how that group was constructed.
  In Top and Mid, we show results for the test point with highest loss; other test points are similar (\refapp{median}), though with more curvature for test loss (\refapp{supp-curvature}).
  The grey reference line has slope 1, and the red borders represent points that are not plotted because they are outside the x- or y-axis range.
  We omit the top row for MNIST, as $\theta^\top \xtest$ is not meaningful in the multi-class setting.
  }
  \vspace{-12mm}
  \label{fig:acc}
\end{figure}

Surprisingly (to us), we found that even when removing large and coherent groups of points, the influence $\predI(w)$ behaved consistently relative to the actual effect $\actI(w)$ on test predictions, test losses, and self-loss, with two broad phenomena emerging:
\begin{enumerate}
  \item {\bf Correlation}: $\predI(w)$ and $\actI(w)$ rank subsets of points $w$ similarly (e.g., high Spearman $\rho$).
  \item {\bf Underestimation}: $\predI(w)$ and $\actI(w)$ tend to have the same sign with $|\predI(w)| < |\actI(w)|$.\footnote{
  This holds with one exception: when measuring the change in test loss,
  $f(\theta) = \ell(\xtest, \ytest; \theta)$,
  underestimation only holds when actual effect $\actI(w)$ is positive (\reffig{acc}-Mid).
  }
\end{enumerate}
Here, we report results on 5 datasets chosen to span a range of applications, training set size $n$, and number of features $d$ (\reftab{data-characteristics}).\footnotemark{}
In an attempt to make the influence approximation as inaccurate as possible,
we constructed a variety of subsets, from small ($\alpha = 0.25\%$) to large ($\alpha = 25\%$), to be coherent and have considerable influence on the model.
On each dataset, we trained an $L_2$-regularized logistic regression model (or softmax for the multiclass tasks)
and compared the influences and actual effects of these subsets.

\footnotetext{
The first 4 datasets involve hospital readmission prediction, spam classification, and object recognition, and were used in \citet{koh2017understanding} to study the influence of individual points. The fifth dataset is a chemical-disease relationship (CDR) dataset \citet{hancock2018babble}.
In \refsec{applications}, we will also study the MultiNLI language inference dataset \citep{williams2018broad}, which was omitted from the experiments here because its large size makes repeated retraining to compute the actual effect too expensive.
See \refapp{ds-details} for dataset details.}

\begin{table*}[t]
\centering
\begin{tabular}{lrrrrrl}
Dataset          & Classes & $n$       & $d$     & $\lambda/n$ & Test acc. & Source \\
\hline
Diabetes         &  $2$    & $20,000$  & $127$   & $2.2\times 10^{-4}$    & $68.2\%$  & \citet{strack2014impact} \\
Enron            &  $2$    & $4,137$   & $3,289$ & $1.0\times 10^{-3}$    & $96.1\%$  & \citet{metsis2006spam} \\
Dogfish          &  $2$    & $1,800$   & $2,048$ & $2.2\times 10^{-2}$    & $98.5\%$  & \citet{koh2017understanding} \\
MNIST            &  $10$   & $55,000$  & $784$   & $1.0\times 10^{-3}$    & $92.1\%$  & \citet{lecun1998gradient} \\
CDR              &  $2$    & $24,177$  & $328$   & $1.0 \times 10^{-4}$    & $67.4\%$  & \citet{hancock2018babble} \\
MultiNLI         &  $3$    & $392,702$ & $600$     & $1.0 \times 10^{-4}$    & $50.4\%$  & \citet{williams2018broad}
\end{tabular}
\caption{Dataset characteristics and the test accuracies that logistic regression achieves (with regularization $\lambda$ selected by cross-validation). $n$ is the training set size and $d$ is the number of features.}
\label{tab:data-characteristics}
\end{table*}

\paragraph{Group construction.}
Our aim is to construct coherent groups that when removed will substantially change the model. To do so, we need to choose points that are similar in some way.
Specifically, for each dataset, we grouped points in 7 ways: 1) points that share feature values;
2) points that cluster on their features or 3) on their gradients $\nt\ell(x, y, \hto)$);
4) random points within the same class; 5) random points from any class.
We also grouped
6) points with large positive and 7) negative influence on the test loss $\ell(\xtest, \ytest, \hto)$,
since intuitively, training points that all have high influence on a test point should act together to change the model substantially.
Overall, for each dataset, we constructed
1,700 subsets ranging in size from $0.25\%$ to $25\%$ of the training points. See \refapp{expt-details} for more details.

\vspace{-1mm}
\paragraph{Results.} \reffig{acc} shows that the influences and actual effects of all of these subsets on test prediction (Top), test loss (Mid), and self-loss (Bot) are highly correlated (Spearman $\rho$ of $0.89$ to $0.99$ across all plots),
even though the absolute and relative errors of the influence approximation can be quite large.
Moreover, the influence of a group tends to underestimate its actual effect in all settings except for groups with negative influence on test loss (the left side of each plot in \reffig{acc}-Mid).
These trends held across a wide range of regularizations $\lambda$, though correlation increased with $\lambda$ (\refapp{reg}).

In \refsec{applications}, we will use the CDR dataset \citep{hancock2018babble} and the MultiNLI \citep{williams2018broad} dataset to show that correlation and underestimation also apply to groups of data that arise naturally, and that influence functions can therefore be used to derive insights about real datasets and applications.
Before that, we first attempt to develop some theoretical insight into the results above.

\section{Theoretical analysis}\label{sec:theory}

The experimental results above show that there is consistent underestimation and high correlation between the predicted effects, based on influence functions, and the actual effects of groups across a variety of datasets, despite the influence approximation incurring large absolute and relative error.
As we discussed in \refsec{related}, this is outside the regime of existing theory.

As an initial step towards understanding the high-error regime,
we establish conditions under which the actual effect $\actI(w)$ lies approximately between $\predI(w)$ and $\Cmax \predI(w)$ for some $\Cmax > 0$.
This cone constraint---so called because it implies that all points on the graph of influence vs. actual effect lie within a cone---implies underestimation and,
if $\Cmax$ is small, some degree of correlation.
We first show that this constraint holds in restricted settings---when measuring self-loss, or when removing multiple copies of the same point---and that $\Cmax$ varies inversely with the regularization term $\lambda$, which is expected since stronger regularization reduces the change in the model.
However, the cone constraint is stronger than necessary because it bounds the degree of underestimation,
and we construct counterexamples to show that it need not hold in more general settings.

Our analysis centers on the \emph{one-step Newton approximation}, which estimates the change in parameters
\begin{align}
  \htow - \hto &\approx \dnw \nonumber
  \eqdef \bigl(\Ho(\ow)\bigr)^{-1} g_\bone(w),
\end{align}
where $\Ho(\ow) = (\sum_{i=1}^n (1 - w_i) \nt^2\ell(x_i, y_i; \hto)) + \lambda I$ is the
regularized empirical Hessian at $\hto$ but reweighted after removing the subset $w$.
This change in parameters gives the Newton approximation of the effect
$\nI(w) = f\bigl(\hto + \dnw\bigr) - f(\hto))$ and the corresponding Newton error $\Errn(w) = \actI(w) - \nI(w)$, which measures its gap from the actual effect. Specifically, we decompose the error between the actual effect $\actI(w)$ and influence $\predI(w)$ as
\begin{align}
  \actI(w) - \predI(w) \ =\  \underbrace{\actI(w) - \nI(w)}_{\Errn(w)} \ +\
  \underbrace{\nI(w) - \predI(w).}_{\Errni(w)}\label{eqn:decomp}
\end{align}
In \refsec{theory-newton}, we first show that the Newton-actual error $\Errn(w)$
decays at a rate of $O\bigl(1 / (\lmin + \lambda)^3\bigr)$, where $\lambda$ is regularization strength and $\lmin$ is the smallest eigenvalue of the empirical Hessian $H_\bone$. Empirically, this error is small on our datasets, so we focus on characterizing the Newton-influence error $\Errni(w)$ in \refsec{theory-decompose}.
We use this characterization to study the behavior of influence relative to the actual effect on self-loss (\refsec{theory-selfloss}) and test prediction (\refsec{theory-test}). For margin-based models, the test loss is a monotone function of the test prediction, so the analysis is similar (\refapp{theory-loss}).

\subsection{Bounding the error of the one-step Newton approximation}\label{sec:theory-newton}

\begin{figure}[t]
  \centering
  \includegraphics[width=\textwidth]{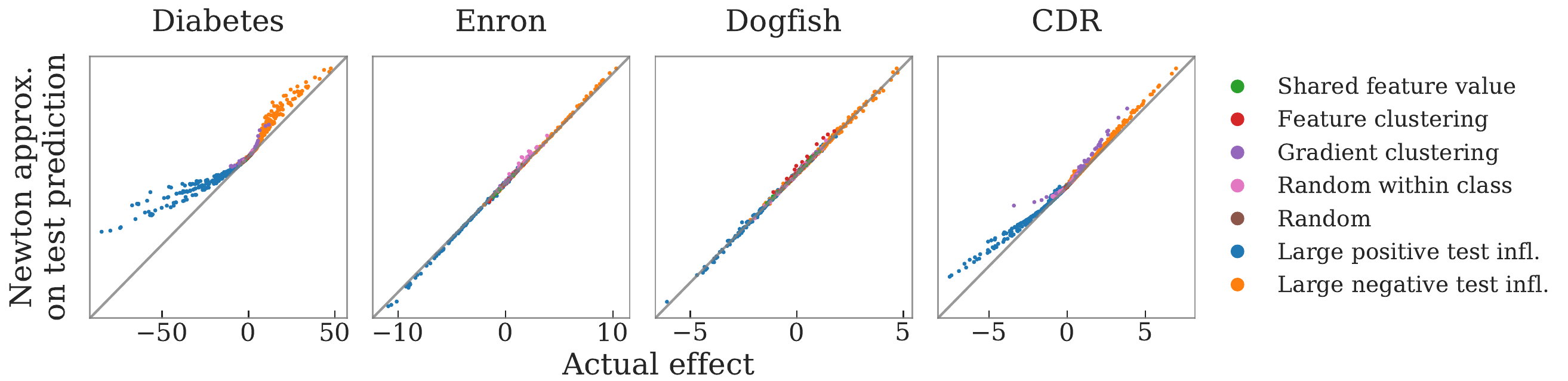}
  \caption{
    The Newton approximation accurately captures the actual effect for our datasets (though there is more error on the Diabetes dataset),
    with the same test point as in \reffig{acc}-Top.
    We omit MNIST and MultiNLI for computational reasons. See \reffig{pparam} for plots of test loss and self-loss.
  }
  \label{fig:newton}
\end{figure}

The Newton approximation is computationally expensive
because it computes $(\Ho(\ow))^{-1}$ for each $w$
(instead of the fixed $\Ho^{-1}$ in the influence calculation).
However, it provides more accurate estimates (e.g., \citet{pregibon1981logistic}, \citet{rad2018scalable}),
and we show that its error can be bounded as follows (all proofs in \refapp{proofs}):
\begin{proposition}\label{prop:newton-actual}
  Let the Newton error be $\Errn(w) \eqdef \actI(w) - \nI(w)$.
  Assume that the evaluation function $f(\theta)$ is $C_f$-Lipschitz
  and that the Hessian $\nt^2 \ell(x, y, \theta)$ is $C_H$-Lipschitz.
  Then
  \begin{align*}
      \abs{\Errn(w)} \leq \frac{n \|w\|_1^2 C_f C_H C_\ell^2}{(\lmin + \lambda)^3},
  \end{align*}
  where we define
  $C_\ell \eqdef \max_{1 \leq i \leq n} \|\nabla_\theta \ell(x_i,y_i,\hto)\|_2$ to be the largest norm of a training point's gradient at $\hto$,
  and $\lmin$ to be the smallest eigenvalue of $H_\bone$.
  $\Errn(w)$ only involves third-order or higher derivatives of the loss, so it is 0 for quadratic losses.
\end{proposition}
\refprop{newton-actual} tells us that the Newton approximation is accurate
when $\lambda$ is large or the third derivative of $\ell(x, y; \cdot)$ (controlled by $C_H$) is small. Empirically,
the Newton error $\Errn(w)$ is strikingly small in most of our settings (\reffig{newton}), even though the overall error of the influence approximation $\actI(w) - \predI(w)$ is still large.
In the remainder of this section, we therefore focus on characterizing the Newton-influence error $\Errni(w)$, under the assumption that the Newton approximation is similar to the actual effect (within a factor of $O(1/\lambda^3)$).

\subsection{Characterizing the difference between the Newton approximation and influence}\label{sec:theory-decompose}
We next characterize the Newton-influence error $\Errni(w) = \nI(w) - \predI(w)$:
\begin{proposition}\label{prop:newton-inf}
  Under the assumptions of \refprop{newton-actual} and the additional assumption that the third derivative of $f(\theta)$ exists and is bounded in norm by $C_{f, 3}$, the Newton-influence error $\Errni(w)$ is
  \begin{align*}
    \Errni(w) &= \nt f(\hto)^\top \Ho^{-\inv{2}} D(w) \Ho^{-\inv{2}} g_\bone(w)
    \ + \ \underbrace{\inv{2} \dnw^\top \nt^2 f(\hto) \dnw + \Errf(w),}_\text{
      Error from curvature of $f(\cdot)$
    }
  \end{align*}
  with $D(w) \eqdef \bigl(I - \Ho^{-\inv{2}}H_\bone(w)\Ho^{-\inv{2}}\bigr)^{-1} - I$
  and $H_\bone(w) \eqdef \sum_{i=1}^n w_i \nt^2\ell(x_i, y_i; \hto)$.
  The \emph{error matrix} $D(w)$ has eigenvalues between 0 and $\frac{\lmax}{\lambda}$,
  where $\lmax$ is the largest eigenvalue of $H_\bone$.
  The residual term $\Errf(w)$ captures the error due to third-order derivatives of $f(\cdot)$ and is bounded by $\abs{\Errf(w)} \leq {\|w\|_1^3 C_{f, 3}C_\ell^3}/{6(\lmin + \lambda)^3}$.
\end{proposition}
We can interpret \refprop{newton-inf} as a formalization of \citet{hampel1986robust}'s observation that influence approximations are accurate when the model is robust and the curvature of the loss is low.
In general, the error decreases as $\lambda$ increases and $f(\cdot)$ becomes less curved;
in \reffig{regularization}, we show that increasing $\lambda$ reduces error and increases correlation in our experiments.

\subsection{The relationship between influence and actual effect on self-loss}\label{sec:theory-selfloss}
Let us now apply \refprop{newton-inf} to analyze the behavior of influence under different choices of evaluation function $f(\cdot)$. We start with the self-loss $f(\theta) = \sum_{i=1}^n w_i \ell(x_i, y_i; \theta)$, as its influences and actual effects are always non-negative, and it is the cleanest to characterize:
\begin{proposition}\label{prop:selfloss}
  Under the assumptions of \refprop{newton-inf}, the influence on the self-loss obeys
  \begin{align*}
    \predI(w) + \Errf(w) &\leq \nI(w)
    \leq \left(1 + \frac{3\lmax}{2\lambda} + \frac{\lmax^2}{2\lambda^2}\right) \predI(w) + \Errf(w).
  \end{align*}
\end{proposition}
The constraint in \refprop{selfloss} implies that up to $O(1/\lambda^3)$ terms, influence underestimates the Newton approximation and therefore the actual effect.
This explains the previously-unexplained downward bias observed when using influence to approximate LOOCV \citep{debruyne2008model, giordano2019swiss}.
Equivalently, all points on the graph of influences vs. actual effects
lie within the cone bounded by the lines with slope $1$ and slope $\frac{\lambda}{\lambda + 3\lmax/2}$ lines, up to $O(1/\lambda^3)$ terms.
As $\lambda$ grows, these lines will converge, and the error terms $\Errf(w)$ and $\Errn(w)$ will decay at a rate of $O(1/\lambda^3)$,
forcing the influences and actual effects to be equal.

However, $\lambda / \lmax$ is quite small in our experiments in \refsec{accuracy},
so the actual correlation of influence is better than predicted by this theory:
in \reffig{acc}-Bot, the sizes of the theoretically-permissible cones can be quite large, but the points in the graphs nevertheless trace a tight curve through the cone.

\subsection{The relationship between influence and actual effect on a test point}\label{sec:theory-test}

\begin{figure}[t]
  \centering
  \floatbox[{\capbeside\thisfloatsetup{capbesideposition={right,top},capbesidewidth=6.5cm}}]{figure}[\FBwidth]
  {\caption{
    Influence $\predI(w)$ vs. Newton approximation $\nI(w)$ on the test prediction on two  counterexamples detailed in \refapp{theory-cex}.
    Left: We adversarially choose a set of $w$'s such that $\predI(w)$ and $\nI(w)$ can have different signs and need not correlate.
    Right: When we only remove copies of single points, underestimation holds. However, we can control the scaling factor $d(w)$ between $\predI(w)$ and $\nI(w)$ on different groups, so correlation need not hold.
  }\label{fig:cex}}
  {\includegraphics[width=0.5\textwidth]{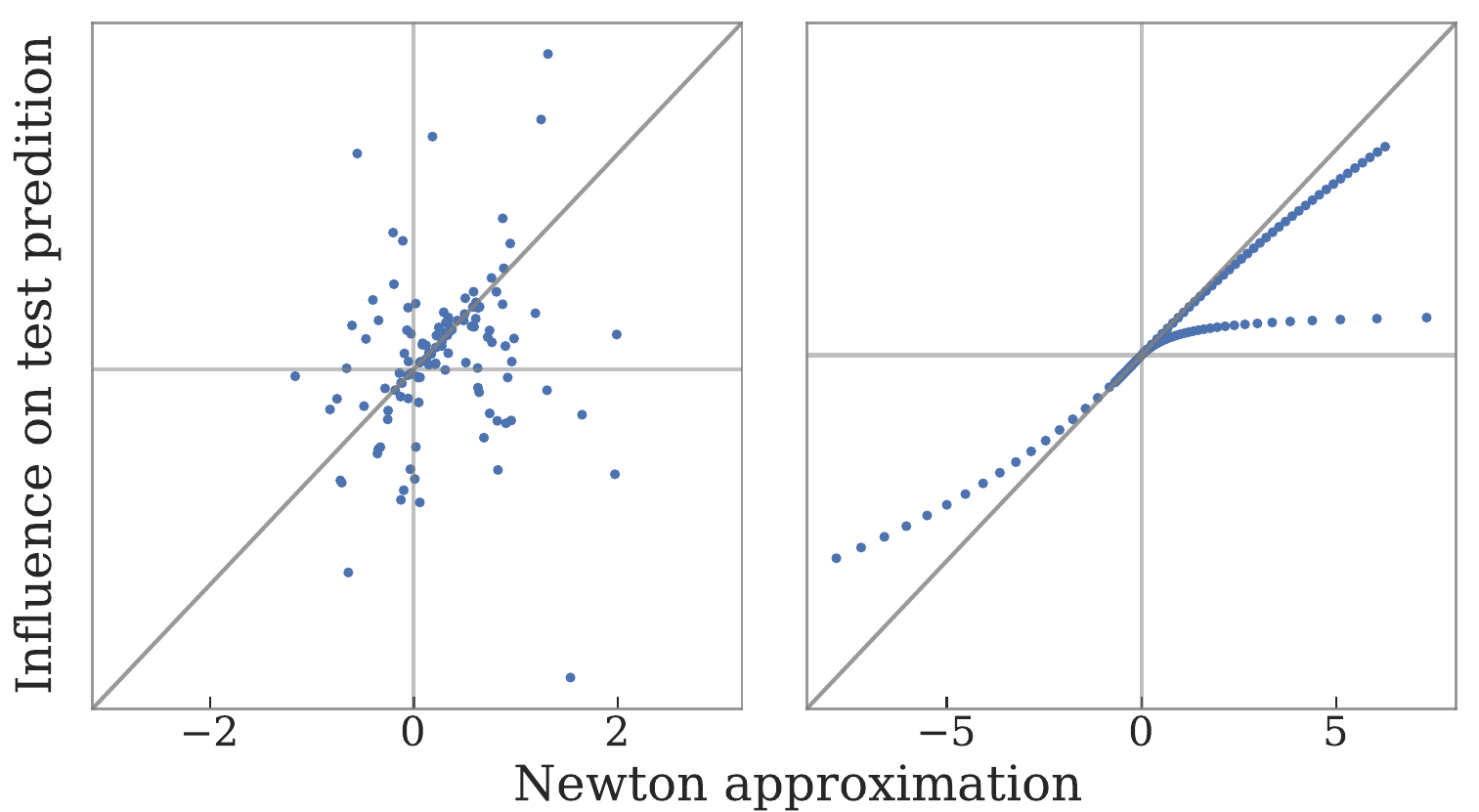}}
\end{figure}

We now turn to measuring the test prediction $f(\theta) = \theta^\top \xtest$. Here, we show that correlation and underestimation need not hold, and that we cannot obtain a cone constraint similar to \refprop{selfloss} except in a restricted setting.
Define $\vtest = \Ho^{-\inv{2}} \xtest$ and $v_w = \Ho^{-\inv{2}} g_\bone(w)$.
\refprop{newton-inf} gives:
\begin{corollary}\label{cor:xtest}
  Suppose $f(\theta) = \theta^\top \xtest$.
  Then $\nI(w) = \predI(w) + \vtest^\top D(w) v_w$,
  where $D(w) = \bigl(I - \Ho^{-\inv{2}}H_\bone(w)\Ho^{-\inv{2}}\bigr)^{-1} - I$ is the \emph{error matrix} from \refprop{newton-inf}.
\end{corollary}
Unfortunately, \refcor{xtest} implies that no cone constraint applies: in general, we can find $\xtest$ such that the influence $\predI(w) = \vtest^\top v_w = 0$ but the Newton approximation $\nI(w) = \vtest^\top D(w) v_w$ is large.
As a counterexample, \reffig{cex}-Left shows that on synthetic data, $\predI(w)$ and $\nI(w)$ can even have opposite signs on some subsets $w$.

We can recover a cone constraint similar to \refprop{selfloss} if we restrict our attention to
the special case where we use a margin-based model and remove (possibly multiple copies) of a single point:
\begin{proposition}\label{prop:single}
  Consider a binary classification setting with $y \in \{-1, +1\}$ and a margin-based model with loss
    $\ell(x, y; \theta) = \phi(y\theta^\top x)$ for some $\phi:\R\to\R_+$.
    Suppose $f(\theta) = \theta^\top \xtest$
    and that the subset $w$ comprises $\|w\|_1$ identical copies of the training point $(x_w, y_w)$.
    Then under the assumptions of \refprop{newton-actual},
    the Newton approximation $\nI(w)$ is related to the influence $\predI(w)$ according to
  \begin{equation*}
    \nI(w) = \frac{\predI(w)}
        {1 - \onenorm{w} \cdot \phi''(y_w \hto^\top x_w) \cdot x_w^\top \Ho^{-1} x_w}.
  \end{equation*}
  This implies the Newton approximation $\nI(w)$ is bounded between $\predI(w)$ and
  $\bigl(1 + \frac{\lmax}{\lambda}\bigr) \predI(w)$.
\end{proposition}
Similar to \refprop{selfloss}, \refprop{single} shows that up to $O(1/\lambda^3)$ terms, the influence underestimates the actual effect when removing copies of a single point.
Moreover, all points on the graph of influences vs. actual effects lie within the cone bounded by the lines with slope $1$ and slope $\lambda / (\lambda + \lmax)$,
up to $O(1/\lambda^3)$ terms.
As $\lambda / \lmax$ grows, the cone shrinks, and correlation increases.

However, if $\lambda / \lmax$ is small (as in our experiments in \refsec{accuracy}),
the cone is wide,
and the scaling factor $d(w) = 1 / (1 - \onenorm{w} \cdot \phi_k'' x_k^\top \Ho^{-1} x_k)$ in \refprop{single} can be quite large for some subsets $w$ but not for others.
In particular, $d(w)$ is large when there are few remaining points in the direction of the removed points.
In \reffig{cex}-Right, we exploit this fact to show that the influence $\predI(w)$ and Newton approximation $\nI(w)$ can exhibit low correlation (e.g., low $\predI(w)$ need not mean low $\nI(w)$),
even in the simplified setting of removing copies of single points.
We comment on the analogue of $d(w)$ in the general multiple-point setting in \refapp{theory-D}, and on the influence on test loss (instead of test prediction) in \refapp{theory-loss}.

\section{Applications of influence functions on natural groups of data}\label{sec:applications}

The analysis in \refsec{theory} shows that
the cone constraint between predicted and actual group effects need not always hold.
Nonetheless, our experiments in \refsec{accuracy} demonstrate that on real datasets, the correlation is much stronger than the theory predicts.
We now turn to using influence functions to predict group effects in two case studies where groups arise naturally.

\paragraph{Chemical-disease relation (CDR).}
The CDR dataset tackles the following task: given text about the relationship between a chemical and a disease, predict if the chemical causes the disease.
It was collected via data programming, where users provide labeling functions (LFs)---instead of labels---that take in an unlabeled point and either abstain or output a heuristic label \citep{ratner2016data}. Specifically, \citet{hancock2018babble} collected natural language explanations of provided classifications; parsed those explanations into LFs; and used those LFs to label a large pool of data (\refapp{bl-details}).

\begin{figure}[t]
  \centering
  \vspace{-4mm}
  \floatbox[{\capbeside\thisfloatsetup{capbesideposition={right,top},capbesidewidth=58mm}}]{figure}[\FBwidth]
  {\caption{
    In CDR, the influence of a labeling function (LF) on test performance is predicted by its coverage (Left) but not its precision (Mid).
    However, in MultiNLI, the number of examples contributed by a crowdworkers is not predictive of its influence (Right).
    For CDR, LFs output either all $+$ or all $-$ labels; we plot the influence of each LF on the test points of the same class.
  }\label{fig:combined-apps}}
  {\includegraphics[width=0.55\textwidth]{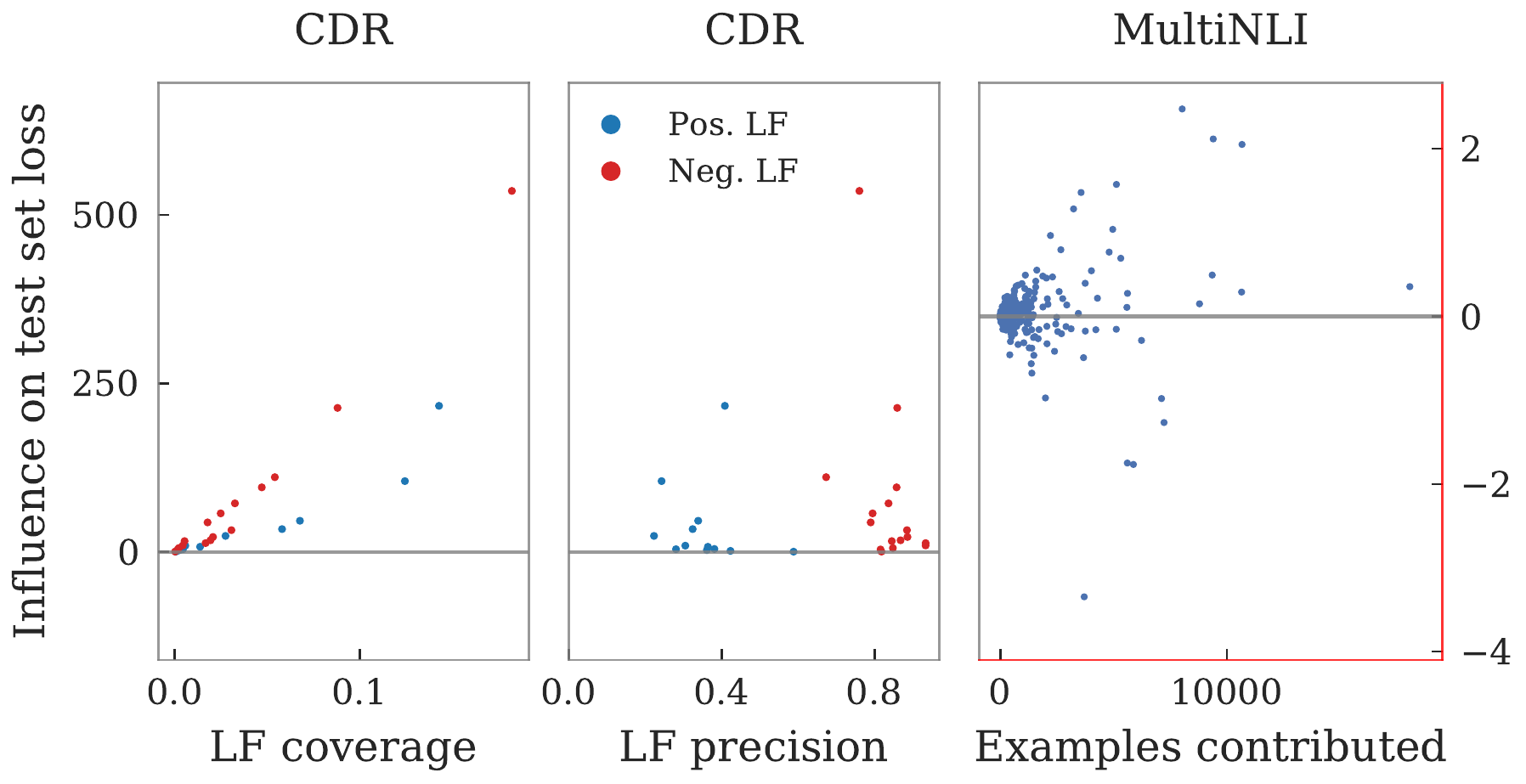}}
\end{figure}

We used influence functions to study two important properties of LFs: \emph{coverage}, the fraction of unlabeled points for which an LF outputs a non-abstaining label; and \emph{precision}, the proportion of correct labels output.
We associated each LF with the group of points that it labeled, and computed its influence; as expected, these correlated with actual effects on overall test loss (Spearman $\rho=1$; \reffig{babble-acc}).
LFs with higher coverage had more influence (\reffig{combined-apps}-Left; see also \reffig{babble-size-corr}), but surprisingly, LFs with higher precision did not (\reffig{combined-apps}-Mid).
The association with coverage stems at least partially from class balance:
each LF outputs either all positive or all negative labels,
so removing an LF with high coverage changes the class balance and consequently improves test performance on one class at the expense of the other (\reffig{combined-apps}-Left).
While these findings are not causal claims, they suggest that the coverage of an LF, rather than its precision, might have a stronger effect on its overall contribution to test performance.

\paragraph{MultiNLI.}
The MultiNLI dataset deals with natural language inference: determining if a pair of sentences agree, contradict, or are neutral. \citet{williams2018broad} presented
crowdworkers with initial sentences from five genres and asked them to generate follow-on sentences that were neutral or in agreement/contradiction (\refapp{multinli-details}).
We studied the effect that each crowdworker had on the model's test set performance by computing the influence of the examples they created on overall test loss (Spearman $\rho$ of $0.77$ to $0.86$ with actual effects across different genres; see \reffig{mnli-acc}).

Studying the influence of each crowdworker reveals that the number of examples a crowdworker created was not predictive of influence on test performance: e.g., the most prolific crowdworker contributed 35,000 examples but had negative influence, and we verified that removing all of those examples and retraining the model indeed made overall test performance worse (\reffig{combined-apps}-Right).
Curiously, this effect was genre-specific: crowdworkers who improved performance on some genres would lower performance on others (\reffig{mnli-genre-categories}), even though the number of examples they contributed to a genre did not correlate with their influence on it (\reffig{mnli-genre-contrib}).
We note that these results are obtained on a baseline logistic regression model built on top of a continuous bag-of-words representation.
Identifying precisely what makes a crowdworker's contributions useful, especially on higher-performing models, could help us improve dataset collection and credit attribution as well as better understand the biases due to annotator effects \citep{geva2019annotator}.

\section{Discussion}\label{sec:discussion}

In this paper, we showed empirically that the influences of groups of points are highly correlated with, and consistently underestimate, their actual effects across a range of datasets, types of groups, and sizes.
These phenomena allows us to use influence functions to better understand the ``different stories that different parts of the data tell,'' in the words of \citet{hampel1986robust}.
We showed that we can gain insight into the effects of a labeling function in data programming, or a crowdworker in a crowdsourced dataset, by computing the influence of their corresponding group effects.

While these applications involved predefined groups, influence functions could potentially also discover coherent, semantically-relevant groups in the data.
They can also be used to approximate Shapley values,
which are a different but related way of measuring the effect of data points;
see, e.g., \citet{jia2019towards} and \citet{ghorbani2019data}.
Separately, influence functions can also estimate the effects of \emph{adding} training points. In this context, underestimation turns into overestimation, i.e., the influence of adding a group of training points tends to overestimate the actual effect of adding that group. This raises the possibility of using influence functions to evaluate the vulnerability of a given dataset and model to data poisoning attacks \citep{steinhardt2017certified}.

Our theoretical analysis showed that while correlation and underestimation hold in some restricted settings, they need not hold in general, realistic settings. This gap between theory and experiments opens up important directions for future work:
Why do we observe such striking correlation between predicted and actual effects on real data?
To what extent is this due to the specific model, datasets, or subsets used?
Do these trends hold for non-convex models like neural networks?
Our work suggests that there could be distributional assumptions that hold for real data and give rise to the broad phenomena of correlation and underestimation.
One promising lead is the surprising observation that the Newton approximation is much more accurate than influence at predicting group effects,
which holds out the hope that we can understand group effects using just low-order terms (since the Newton approximation only uses the first and second derivatives of the loss) without needing to account for the whole loss function through higher order terms (as in \citet{giordano2019higher}).

\newpage
\subsubsection*{Reproducibility}
The code for replicating our experiments is available in the GitHub repository \url{https://github.com/kohpangwei/group-influence-release}.
An executable version of this paper is also available on CodaLab at \url{https://worksheets.codalab.org/worksheets/0xfed2ae0b9e5b44b7a1af8096365592a5}.

\subsubsection*{Acknowledgments}
We are grateful to
Zhenghao Chen,
Brad Efron,
Jean Feng,
Tatsunori Hashimoto,
Robin Jia,
Stephen Mussmann,
Aditi Raghunathan,
Marco T\'ulio Ribeiro,
Noah Simon,
Jacob Steinhardt,
and Jian Zhang
for helpful discussions and comments.
We are further indebted to
Ryan Giordano,
Ruoxi Jia,
and Will Stephenson for discussion about prior work,
and
Samuel Bowman,
Braden Hancock,
Emma Pierson,
and Pranav Rajpurkar
for their assistance with applications and datasets.
This work was funded by an Open Philanthropy Project Award.
PWK was supported by the Facebook Fellowship Program.

\bibliography{refdb/all}

\newpage
\appendix
\renewcommand\thefigure{\thesection.\arabic{figure}}
\setcounter{figure}{0}
\section{Experimental details for comparing influence vs. actual effects on constructed groups}\label{sec:expt-details}
\subsection{Model training}
For all experiments in \refsec{accuracy}, we trained a logistic regression model (or softmax for multiclass) using
\texttt{sklearn.linear\_model.LogisticRegression.fit},
 fitting the intercept but only applying $L_2$-regularization to the weights.
To choose the regularization strength $\lambda$, we conducted 5-fold
cross-validation across 10 possible values of $\lambda/n$ logarithmically
spaced between $1.0\times 10^{-4}$ and $1.0\times 10^{-1}$, inclusive,
selecting the regularization that yielded the highest cross-validation accuracy
(except on the CDR dataset, where we selected regularization based on
cross-validation F1 score to account for class imbalance as per
\citet{hancock2018babble}'s procedure).

\subsection{Group construction}\label{sec:subsetgen}
For each dataset, we constructed groups of various sizes relative to the entire dataset by considering 100 sizes linearly spaced between $0.25\%$ and $25\%$ of the dataset. For each of these 100 sizes, we constructed one group with each of the following methods:
\begin{enumerate}
  \item Shared features: We selected a single feature uniformly at random and sorted the dataset along this selected feature. Next, we selected an training point uniformly at random. We then constructed a group of size $s$ that consisted of the $s$ unique training points that were closest to the chosen point, as measured by their values in the selected feature. We randomly sampled a feature and initial training point for each different group constructed in this way.

  \item Feature clustering: We clustered the dataset with respect to raw features via \texttt{scipy.cluster.hierarchy.fclusterdata} with \texttt{t} set to $1$, as well as with \texttt{sklearn.cluster.KMeans.fit} with \texttt{n\_clusters} taking on values $4,8,16,32,64,128$. Since hierarchical clustering determines cluster sizes automatically with a principled heuristic and we try a range of values for \texttt{n\_clusters} in $k$-means, this recovers clusters with a large range of sizes. The clustering with $\texttt{n\_clusters}=4$ also guarantees (via the pigeonhole principle) that there is at least one cluster which contains at least $25\%$ of the dataset. From all the clusters that are at least the size of the desired group, we chose one uniformly at random and chose the group uniformly at random and without replacement from the training points in this cluster.

  \item Gradient clustering: We followed the same procedure as ``Feature clustering,'' except that we clustered the dataset with respect to $\nt \ell(x,y;\hto)$, i.e. each training point was represented by the gradient of the loss on that point.

  \item Random within class: We considered all classes with at least as many training points as the size of the desired group. From these classes, we chose one uniformly at random. Then, we chose the group uniformly at random and without replacement from all training points in this class.

	\item Random: We picked a group uniformly at random and without replacement from the entire dataset.
\end{enumerate}

The above methods gave us a total of 500 groups (100 groups per method) for each dataset, with the exception of the ``random within class'' method for MNIST. Since MNIST has 10 classes, each with only $10\%$ of the data, we skipped over groups of size $>10\%$ just for the ``random within class'' groups.

In addition, we selected 3 random test points and the 3 test points with highest loss; we intend these to represent the average case and the more extreme case that may be relevant to model developers who want to debug errors that their model outputs. For each of these 6 test points, we selected groups that had large positive influence on its test loss. More specifically, we proceeded in 3 stages:
\begin{enumerate}
	\item We considered 33 group sizes linearly spaced between $0.25\%$ and $2.5\%$ of the dataset, and for any size $s$ out of these 33, we selected a group uniformly at random and without replacement from training points in the top $1.5\times 2.5\%$ of the dataset, ordered according to their influence on the test point of interest.
	\item This was similar to the first stage, but with 33 sizes spaced between $0.25\%$ and $10\%$ and groups chosen from the top $1.5\times 10\%$ of the dataset.
	\item Finally, we considered 34 sizes spaced between $0.25\%$ and $25\%$, with groups chosen from the top $1.5\times 25\%$ of the dataset.
\end{enumerate}
Larger groups tend to have lower average influence than smaller groups, since by necessity, the group must contain points farther from the top.
This multi-stage approach ensured that we would select small groups with both a high average influence and also with a low average influence, so that we could compare them to larger groups and mitigate confounding the group size with its average influence.

Finally, we repeated this last method of group construction for groups with large negative influence on test point loss.

Using these 6 test points, we generated 1,200 groups (100 subsets per group, with 6 test points, and drawing from the positive and negative tails). In total, we therefore generated 1,700 groups per dataset (except MNIST).

\subsection{Comparison of influence and actual effect}\label{sec:compare}

To produce \reffig{acc}, we selected groups as described in \refapp{subsetgen}. We retrained the model once for each group, excluding the group in order to calculate its actual effect. To compute all groups' influences, we first calculated the influence of every individual training point using the procedure of \citet{koh2017understanding}. Then, to compute the influence on test prediction or loss of some group, we simply added the relevant individual influences (in CDR, we weighted these individual influences according to that point's weight; see \refapp{bl-details}). To compute the influence on self-loss of some group, we summed up the gradients of the loss of each training point to compute $g_\bone(w)$, we calculated the inverse Hessian vector product $\Ho^{-1} g_\bone(w)$ and took its dot product with $g_\bone(w)$ (again, we modified this with appropriate weighting for individual points in CDR).

\section{Dataset details}\label{sec:ds-details}
We used the same versions of the Diabetes, Enron, Dogfish, and MNIST datasets as \citet{koh2017understanding}, since the examination of the accuracy of influence functions for large perturbations is a natural extension of their studies of small perturbations. Additionally, we applied influence to more natural settings in CDR and MultiNLI; here, we discuss their preprocessing pipelines.

\subsection{CDR}\label{sec:bl-details}

\citet{hancock2018babble} established the BabbleLabble framework for data programming, following the following pipeline: They took labeled examples with natural language explanations, parsed the explanations into programmatic labeling functions (LFs) via a semantic parser, and filtered out obviously incorrect LFs. Then, they applied the remaining LFs to unlabeled data to create a sparse label matrix, from which they learned a label aggregator that outputs a noisily labeled training set. Finally, they ran $L_2$-regularized logistic regression on a set of basic linguistic features with the noisy labels.

They demonstrated their method on three datasets: Spouse, CDR, and Protein. The Protein dataset was not publicly available, and the vast majority of Spouse was labeled by a single LF, hence we chose to use CDR. This dataset's associated task involved identifying whether, according to a given sentence, a given chemical causes a given disease. For instance, the sentence
``Young women on replacement \textit{estrogens} for ovarian failure after cancer therapy may also have increased risk of \textit{endometrial carcinoma} and should be examined periodically.''
would be labeled True, since it indicates that estrogens may cause endometrial carcinoma \citep{hancock2018babble}.
The sentences and ground truth labels were sourced from the 2015 BioCreative chemical-disease relation dataset \citep{wei2015overview}.

In our application, we began with their 28 LFs and the corresponding label matrix. For simplicity, we did not learn a label aggregator; instead, if an example $x$ was given labels $y_{i_1},y_{i_2},\ldots,y_{i_k}$ by $k$ LFs $i_1,\ldots,i_k$, then we created $k$ copies of $x$, each with weight $1/k$. The subset of points corresponding to LF $i_1$ then included one instance of $x$ with weight $1/k$. This weighting was taken into account in model training as well as in calculations of influence and actual effect. In addition, we used $L_1$-regularization for feature selection, reducing the number of features to 328 while still achieving similar F1 score to \citep{hancock2018babble}; they reported an F1 of 42.3, while we achieved 42.0. After feature selection, we remove the $L_1$-regularization and train a $L_2$-regularized logistic regression model. We assume that the feature selection step is static and not affected by removing groups of data (though in general this assumption is not true); we therefore do not include feature selection in our influence calculations.

We note that in BabbleLabble, a given LF can never output positive on one example but negative on another. Hence, some LFs are positive (unable to output negative and only able to abstain or output positive), while the others are negative (unable to output positive and only able to abstain or output negative).

\subsection{MultiNLI}\label{sec:multinli-details}

\citet{williams2018broad} created the MultiNLI dataset for the task of natural language inference: determining if a pair of sentences agree, contradict, or are neutral. To do so, they presented crowdworkers with initial sentences and asked them to generate follow-on sentences that were neutral or in agreement/contradiction. For example, a crowdworker may be presented with ``Met my first girlfriend that way.'' and write the contradicting sentence ``I didn’t meet my first girlfriend until later.'' \citep{williams2018broad}. Thus, each of the 380 crowdworkers generated a subset of the dataset. We used these subsets in our application of influence.

The training set consisted of 392,702 examples from five genres. The development set consisted of 10,000 ``matched'' examples from the same five genres as the training set, as well as 10,000 ``mismatched'' examples from five new genres. The test set was put on Kaggle as an open competition, hence we do not have its labels and could not use it; therefore, we use the development set as the test set.

The continuous bag-of-words baseline in \citet{williams2018broad} first converted the raw text of each sentence in the pair into a vector by treating the sentence as a continuous bag of words and simply averaging the 300-dimensional GloVe vector embeddings. This converted a pair of sentences into vectors $a,b$. They then concatenated $[a,b,a-b,a \odot b]$ into a 1200D vector, where $a \odot b$ denotes the element-wise product. Finally, they treated this as input to a neural network with three hidden layers and fine-tuned the entire model, including word embeddings (more details in \citep{williams2018broad}).

For our application, we truncated their baseline and just used the concatenation of $a$ and $b$ as the representation for every example. By running logistic regression on this, we achieved test accuracy of $50.4\%$ (vs. their baseline's $64.7\%$; the performance difference comes from the additional dimensions in their vector embeddings and the finetuning through the neural network). Future work could explore influence in the setting of more complex and higher-performing models.

\section{Additional experiments}\label{sec:supp-experiments}

As in \reffig{acc}, in each of the plots below, the grey reference line has slope 1, and the red borders represent points that are not plotted because they are outside the x- or y-axis range.

\subsection{Representative test points}\label{sec:median}

\begin{figure}[h]
  \centering
  \includegraphics[width=\textwidth]{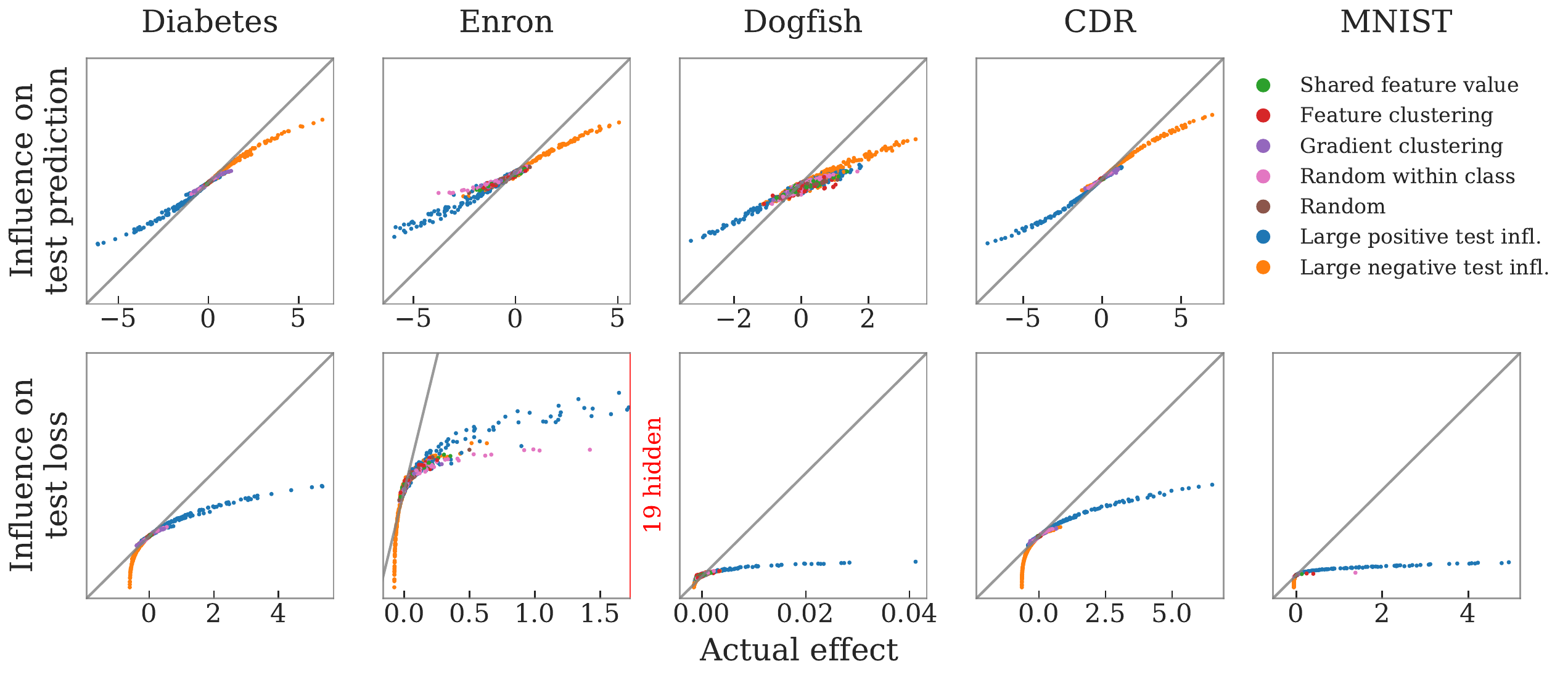}
  \caption{
    Influences vs. actual effects of the same coherent groups in \reffig{acc},
    but on test points closer to the median (within the 40th to 60th
    percentile) of the test loss distribution. We consider these to represent
    average test points. On these, influence on the test prediction remains
    well-correlated with the actual effect.
  }
  \label{fig:median}
\end{figure}

\reffig{median} is similar to \reffig{acc} in the main text:
it shows the influences vs. actual effects of groups on test points,
but with test points that are closer to the median (within the 40th to 60th percentile) of the test loss distribution.

\subsection{Regularization}\label{sec:reg}

\begin{figure}[h!]
  \centering
  \includegraphics[width=0.9\textwidth]{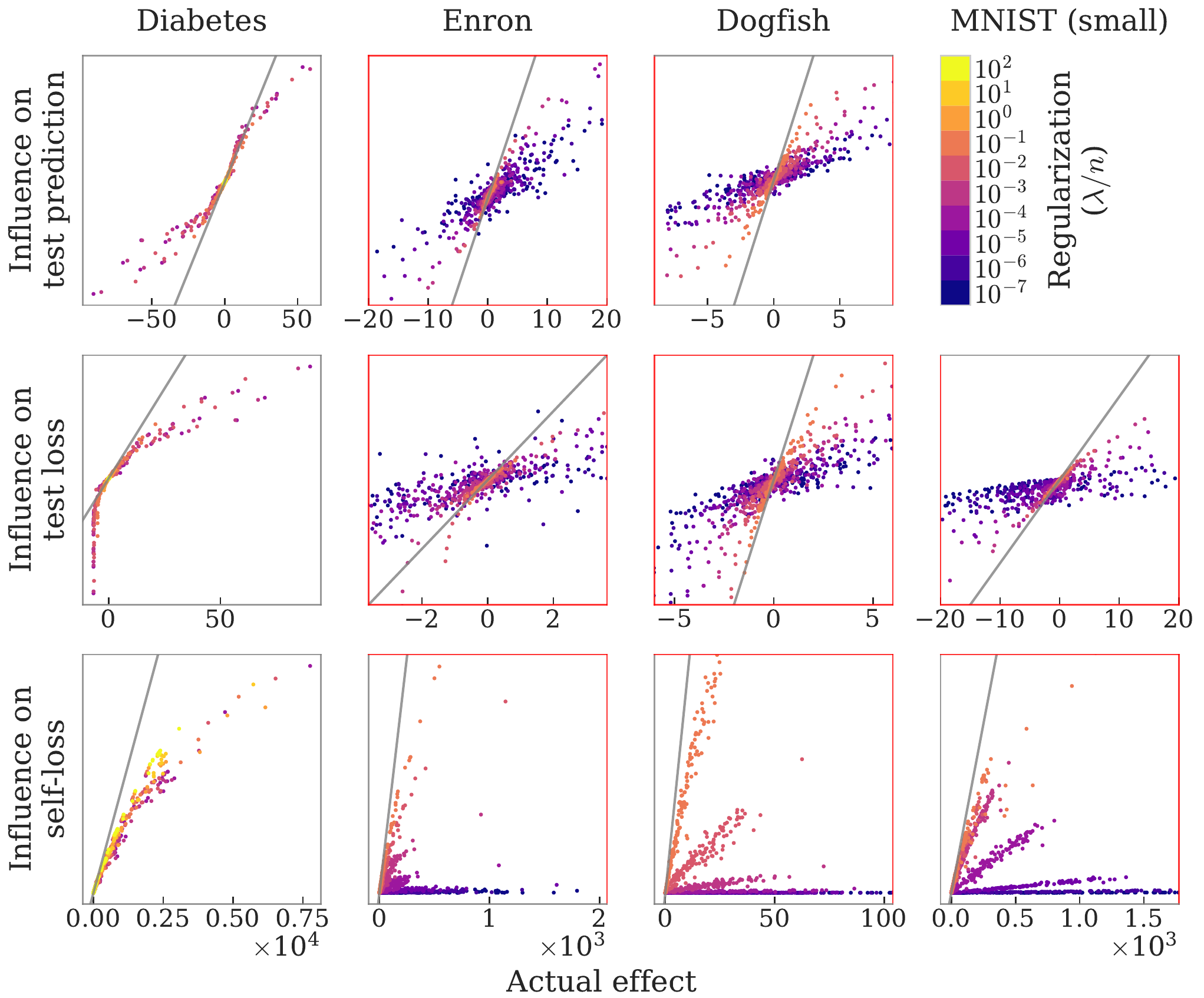}
  \caption{
    The effect of regularization for a representative test point.
    Red frame lines indicate the existence of points exceeding those bounds. We did not include the test prediction for MNIST (small) because the margin is not well-defined for a multiclass model.
  }
  \label{fig:regularization}
\end{figure}

\begin{figure}[h!]
  \centering
  \includegraphics[width=0.95\textwidth]{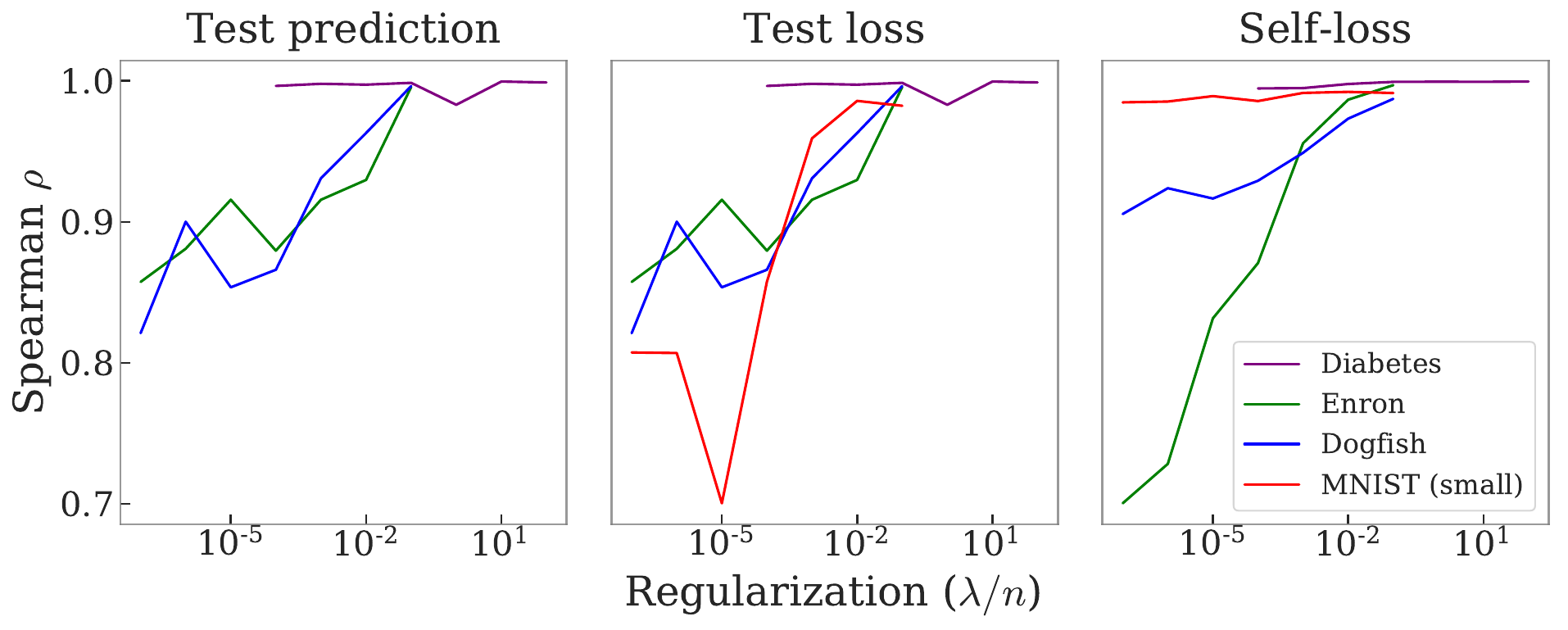}
  \caption{
    As regularization increases, correlation increases between the influences and actual effects on test prediction (Left), test loss (Middle), and self-loss (Right).
  }
  \label{fig:regs-spearman}
\end{figure}

In \refsec{theory}, our bounds show that influence ought to be closer to actual effect as regularization increases. Here, we support this claim empirically on Diabetes, Enron, Dogfish, and MNIST (small).\footnotemark{} To do so, for each dataset, we selected a range of values for $\lambda/n$, and we selected subsets as described in \refapp{subsetgen}. We then computed the influence and actual effect of each of these subsets on a representative test point's prediction, that point's loss, and on self-loss (\reffig{regularization}).

\footnotetext{This experiment required us to retrain the model for every value of $\lambda$ and for every subset. Thus, for computational purposes, we omitted CDR and MultiNLI, and we selected a random $10\%$ subset of MNIST's training set to use in place of all of MNIST.}

In \reffig{regs-spearman}, we observe the trend that correlation generally increases as $\lambda$ does. Specifically, we computed the Spearman $\rho$ between the influence and actual effect for each dataset, each value of $\lambda$, and each evaluation function $f(\cdot)$ of interest (i.e., test prediction, test loss, or self-loss).

\subsection{The effect of loss curvature on the accuracy of influence}\label{sec:supp-curvature}

\begin{figure}[t!]
  \centering
  \includegraphics[width=\textwidth]{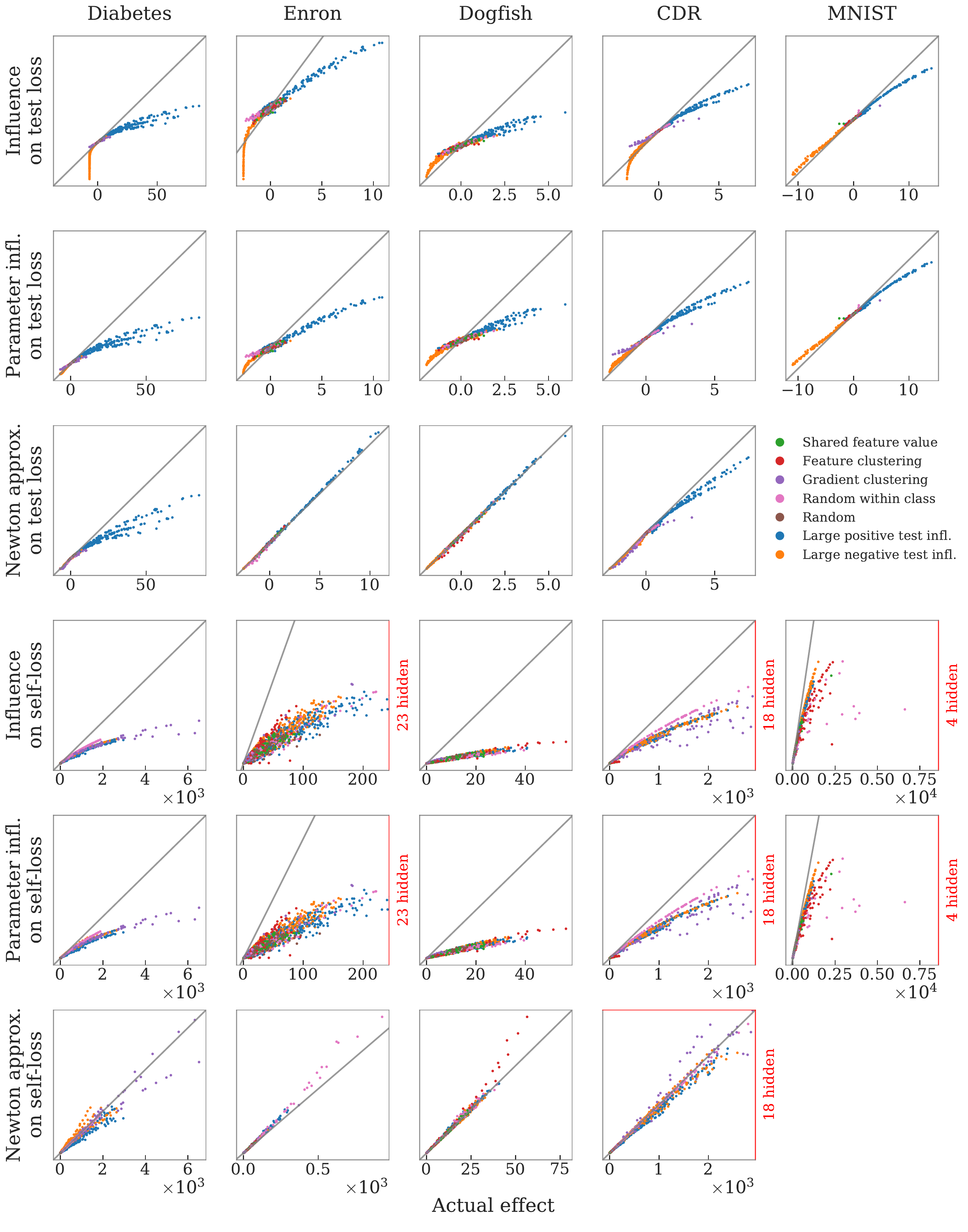}
  \vspace{-4mm}
  \caption{
    The top 3 rows show influence on test loss (with the same test points as in \reffig{acc}),
    while the bottom 3 show self-loss.
    Within each set, the first row shows the influence vs. actual effect (as in \reffig{acc});
    the second shows the predicted effect obtained by estimating the change in parameters via influence and then
    evaluating $f(\cdot)$ directly on those parameters;
    and the third shows the Newton approximation.
    }
    \vspace{-4mm}
  \label{fig:pparam}
\end{figure}

One takeaway from the results on test loss in \reffig{acc}-Mid is that the curvature of $f(\theta)$ can significantly increase
approximation error; this is expected since the influence $\predI(w)$
linearizes $f(\cdot)$ around $\hto$.
When possible, choosing a $f(\cdot)$ that has low curvature (e.g., the linear prediction) will result in higher accuracy.
We can mitigate this by using influence to approximate the parameters $\htow$
and then plug that estimate into $f(\cdot)$ (\reffig{pparam}), though this can be more
computationally expensive.

Note that \reffig{pparam} shows that this technique does not help much for measuring self-loss.
However, in the context of LOOCV,
the computational complexity of the Newton approximation for self-loss (described in \refsec{theory})
is similar to that of the influence approximation,
so we encourage the use of the Newton approximation for LOOCV (as in \citet{rad2018scalable});
\reffig{pparam} shows that this leads to more accurate approximations for self-loss.

\subsection{Additional analysis of influence functions applied to natural groups of data}\label{sec:supp-applications}
In \refsec{applications}, we considered the CDR and MultiNLI datasets, which contain the natural subsets of LFs and crowdworkers, respectively.
To draw inferences about these subsets, we took the $L_2$-regularized logistic regression model described in \refapp{expt-details}, calculated the influence of the LF/crowdworker subsets, and retrained the model once for each LF/crowdworker.

\paragraph{CDR.}
As discussed in \refapp{bl-details}, an LF is either positive or negative, where a positive LF can only give positive labels or abstain, and similarly for negative LFs. Because of this stark class separation, we indicate whether an LF is positive or negative, and we consider LF influence on the positive test examples separately from their influence on the negative test examples. To measure an LF's influence and actual effect on a set of test points, we simply add up its influence and actual effect on the set's individual test points.

\begin{figure}[h]
  \centering
  \includegraphics[width=0.65\textwidth]{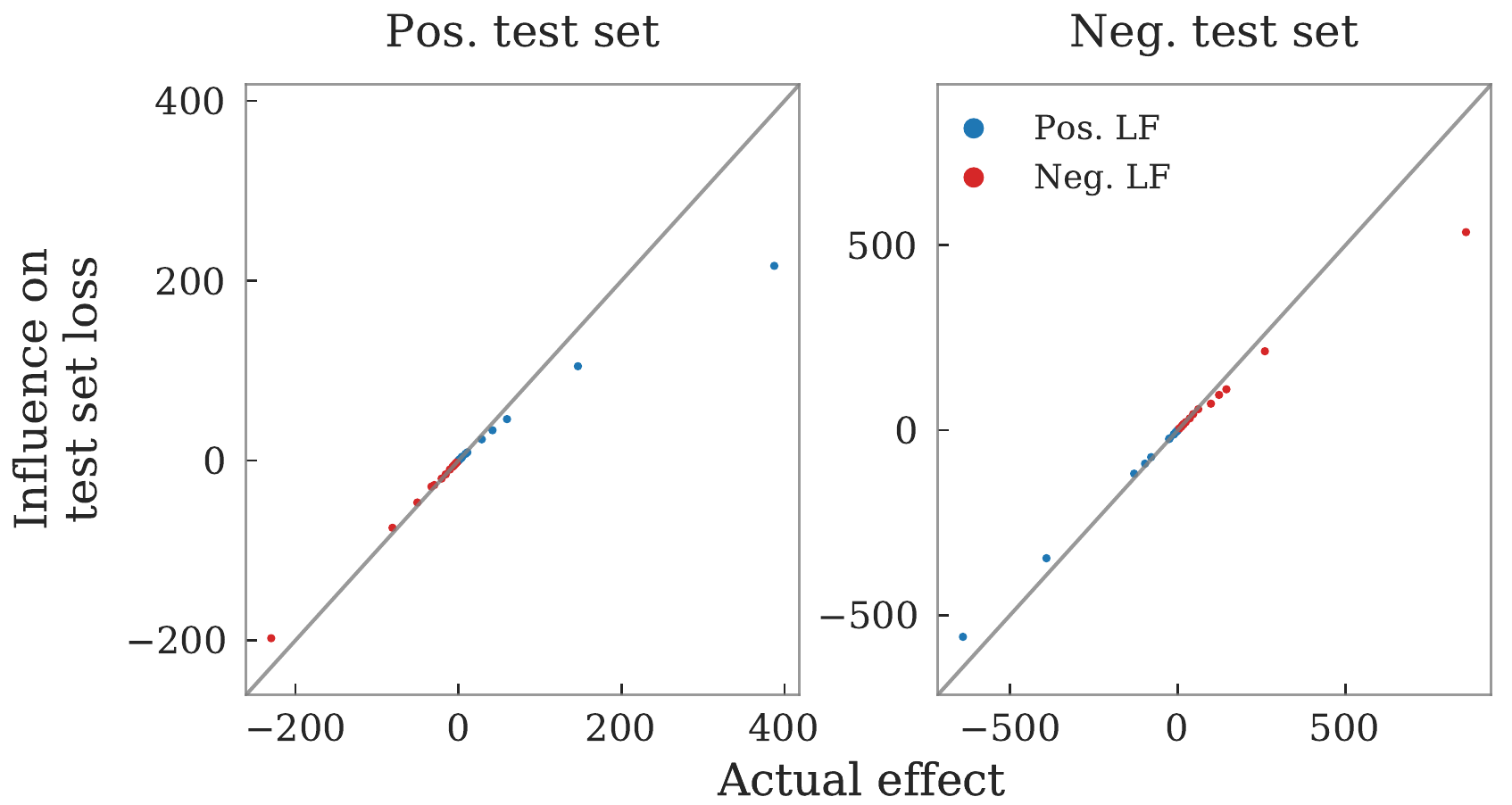}
  \caption{We observe both correlation and underestimation for LFs on the positive and negative test sets. We also see that positive LFs help the positive test set and hurt the negative test set; vice versa for negative LFs.
  }
  \label{fig:babble-acc}
\end{figure}

In \reffig{babble-acc}, we note that influence is a good approximation of an LF's actual effect, just as with other kinds of subsets as well as other datasets (\reffig{acc}). Furthermore, we observe that positive LFs improve the overall performance of the positively labeled portion of the test set while hurting the negatively labeled portion of the test set, and vice versa for negative LFs. This dichotomous effect further motivates the analysis of influence on the positive test set separately from the negative test set, since the process of adding these two influences to study the influence on the entire test set would obscure the full story.

\begin{figure}[h!]
  \centering
  \includegraphics[width=0.65\textwidth]{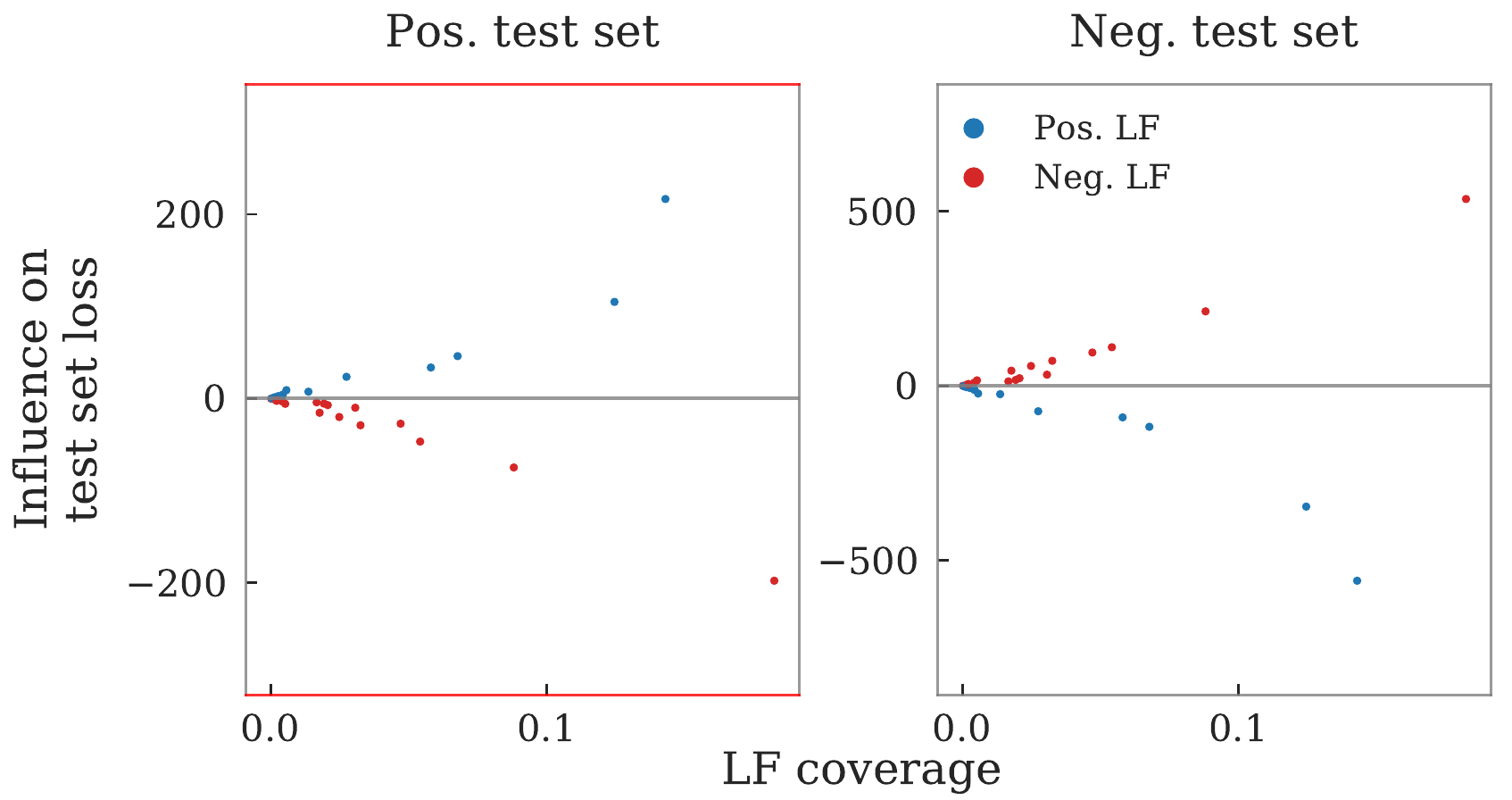}
  \caption{The magnitude of LF influence correlates with coverage. This figure is an extension of \reffig{combined-apps}-Left: there, we showed the influence of positive LFs on the positive test set and the influence of negative LFs on the negative test set. Here, we additionally show the influence of positive LFs on the negative test set and vice versa.
  }
  \label{fig:babble-size-corr}
\end{figure}

Next, we define an LF's \textit{coverage} to be the proportion of the examples that it does not abstain on, which can be measured through the number of examples in its corresponding subset. In \reffig{babble-size-corr}, we observe that the magnitude of influence correlates strongly with coverage.

\begin{figure}[h!]
  \centering
  \includegraphics[width=0.65\textwidth]{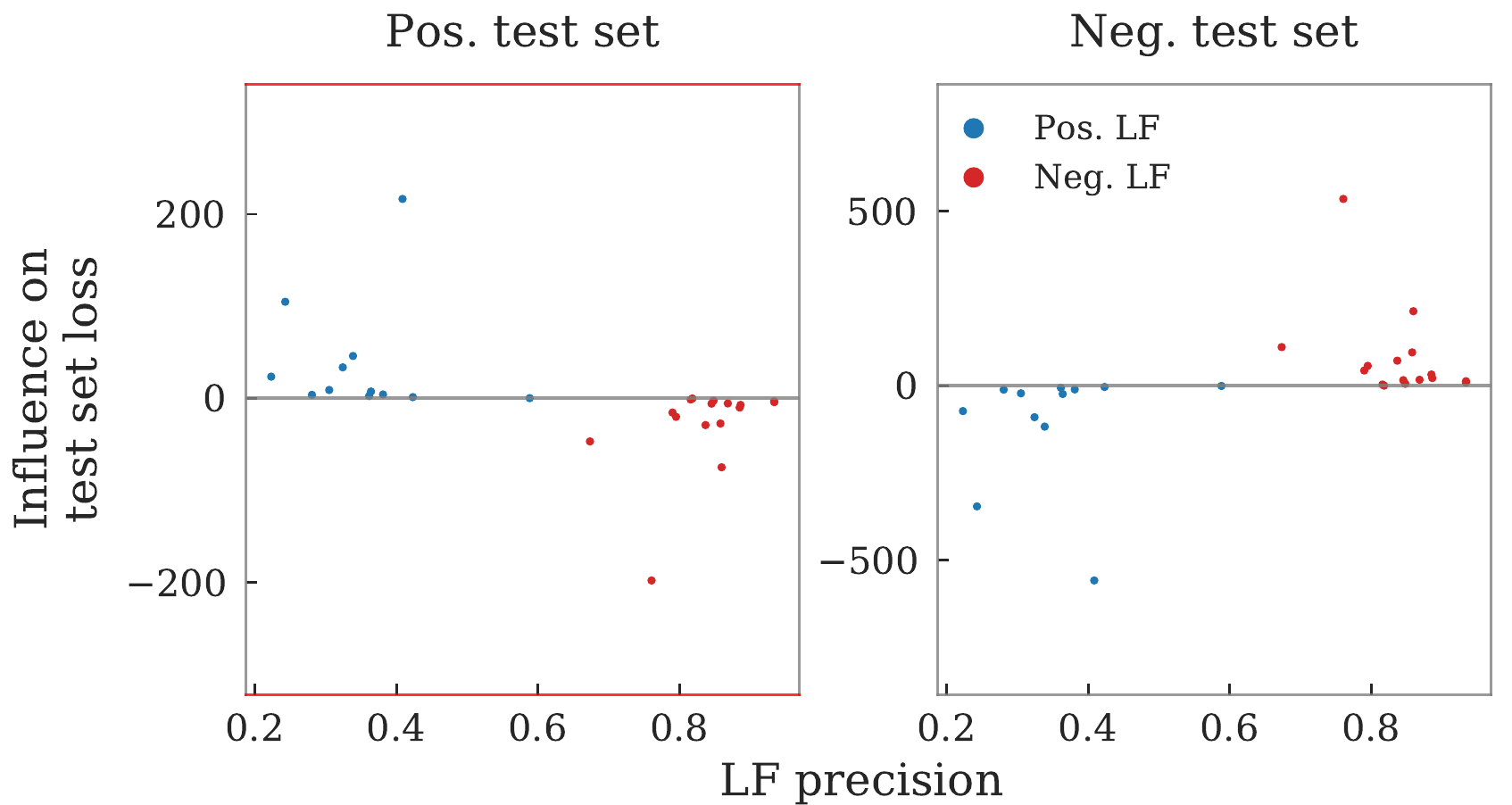}
  \caption{LF influence does not correlate with precision. Similar to \reffig{babble-size-corr}, this figure is an extension of \reffig{combined-apps}-Mid: there, we showed the influence of positive LFs on the positive test set and the influence of negative LFs on the negative test set. Here, we additionally show the influence of positive LFs on the negative test set and vice versa.
  }
  \label{fig:babble-acc-no-corr}
\end{figure}

Finally, we define an LF's \textit{precision} to be the number of examples it labels correctly divided by the number of examples it does not abstain on. Because the dataset had many more negative than positive examples, positive LFs had lower precision than negative LFs. Surprisingly, even when this effect was taken into account and we considered positive LFs separately from negative ones, precision did not correlate with influence (\reffig{babble-acc-no-corr}).

\paragraph{MultiNLI.}
As discussed in \refapp{multinli-details}, the training set consisted of five genres, and the test set consisted of a matched portion with the same five genres, as well as a mismatched portion with five new genres. For succinctness, we refer to the influence/actual effect of the set of examples generated by a single crowdworker as that crowdworker's influence/actual effect.

\begin{figure}[h!]
  \centering
  \includegraphics[width=\textwidth]{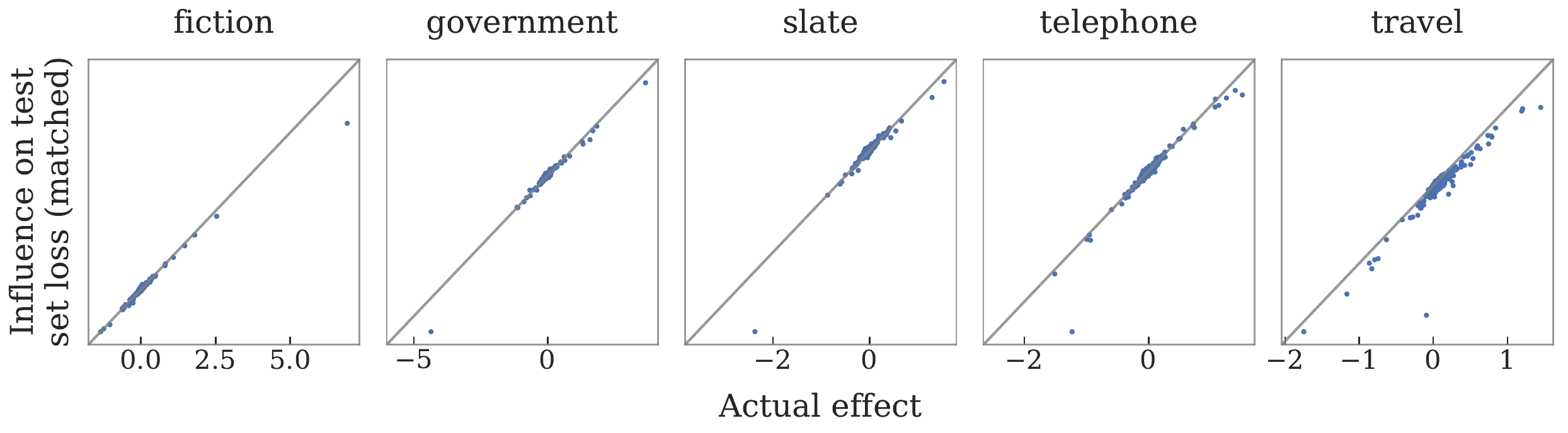}
  \includegraphics[width=\textwidth]{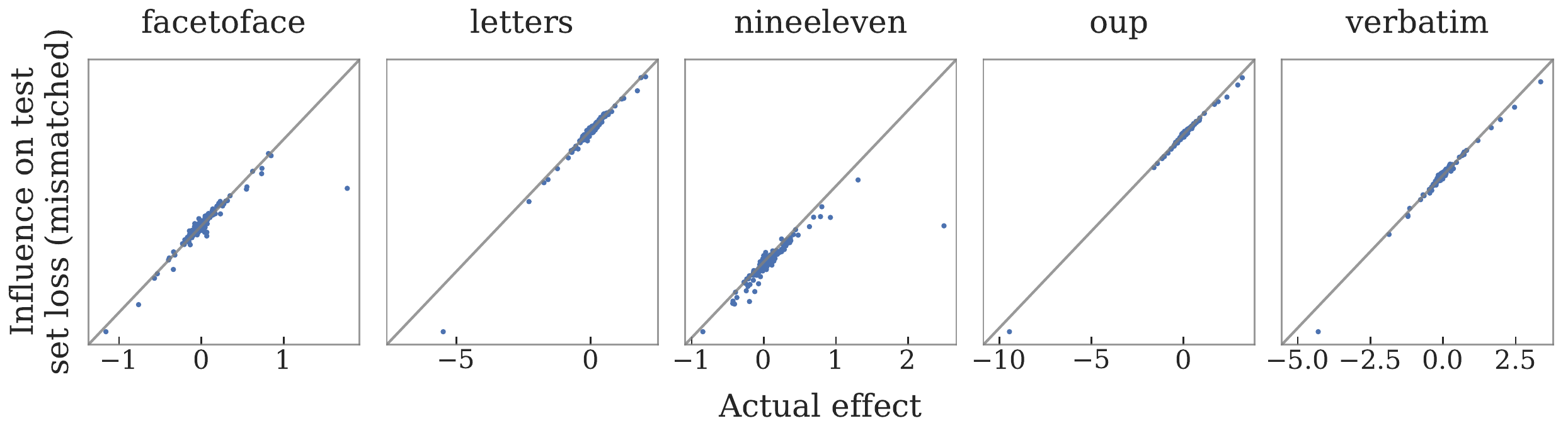}
  \caption{We observe strong correlation between crowdworkers' influence and actual effects.
  }
  \label{fig:mnli-acc}
\end{figure}

First, we note in \reffig{mnli-acc} that influence is a good approximation of a crowdworker's actual effect for both matched and mismatched test sets, consistent with our findings in \reffig{acc} for other subset types and datasets.

\begin{figure}[h!]
  \centering
  \includegraphics[width=0.42\textwidth]{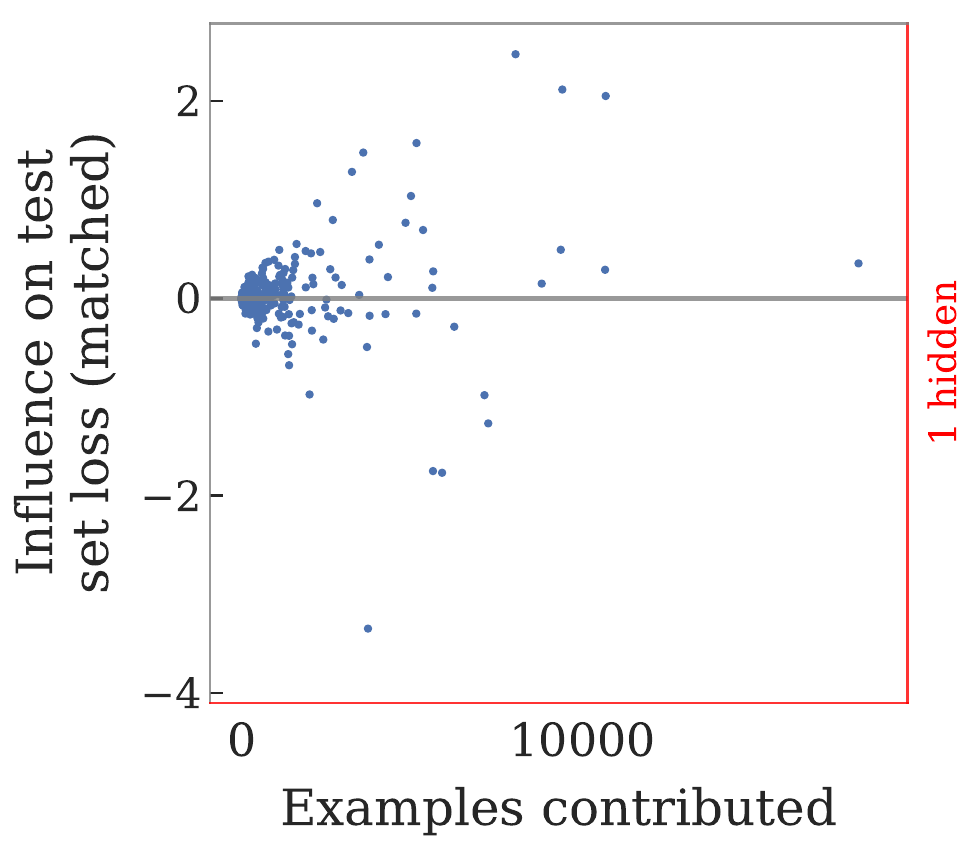}
  \caption{Size does not correlate strongly with influence. The hidden point is the crowdworker that contributed 35,000 examples. This is the figure presented in \reffig{combined-apps}-Right.
  }
  \label{fig:mnli-size-no-corr}
\end{figure}

Unlike in CDR (\reffig{babble-size-corr}), we do not find strong correlation between a crowdworker's influence and the number of examples they contributed; it is possible to contribute many examples but have relatively little influence (\reffig{mnli-size-no-corr}).

\begin{figure}[h!]
  \centering
  \includegraphics[width=0.55\textwidth]{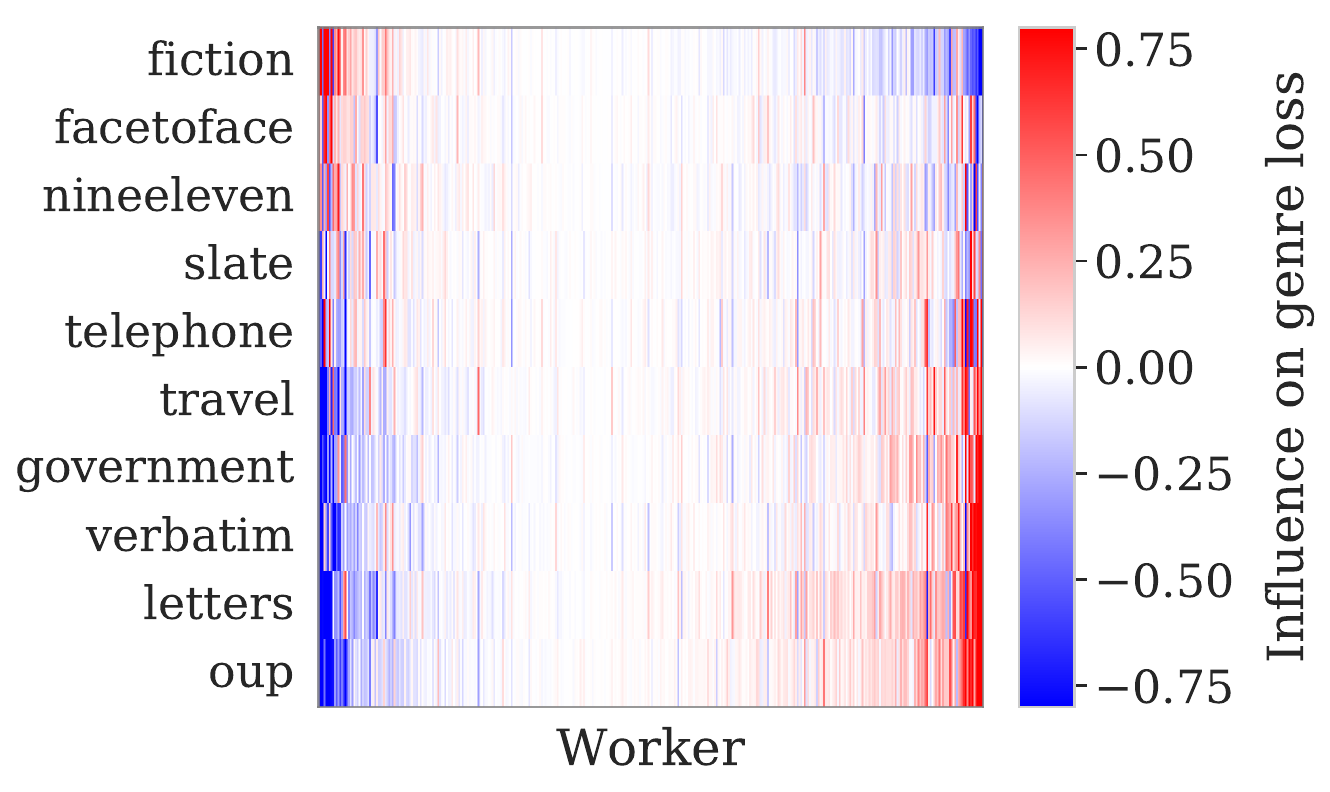}
  \caption{Workers tended to have positive influence on fiction, facetoface, and nineeleven and negative influence on travel, government, verbatim, letters, and oup (or vice versa). In this plot, we allowed for full color saturation when the magnitude of the total influence on the test set (matched) exceeded 0.8.
  }
  \label{fig:mnli-genre-categories}
\end{figure}

\begin{figure}[h!]
  \centering
  \includegraphics[width=\textwidth]{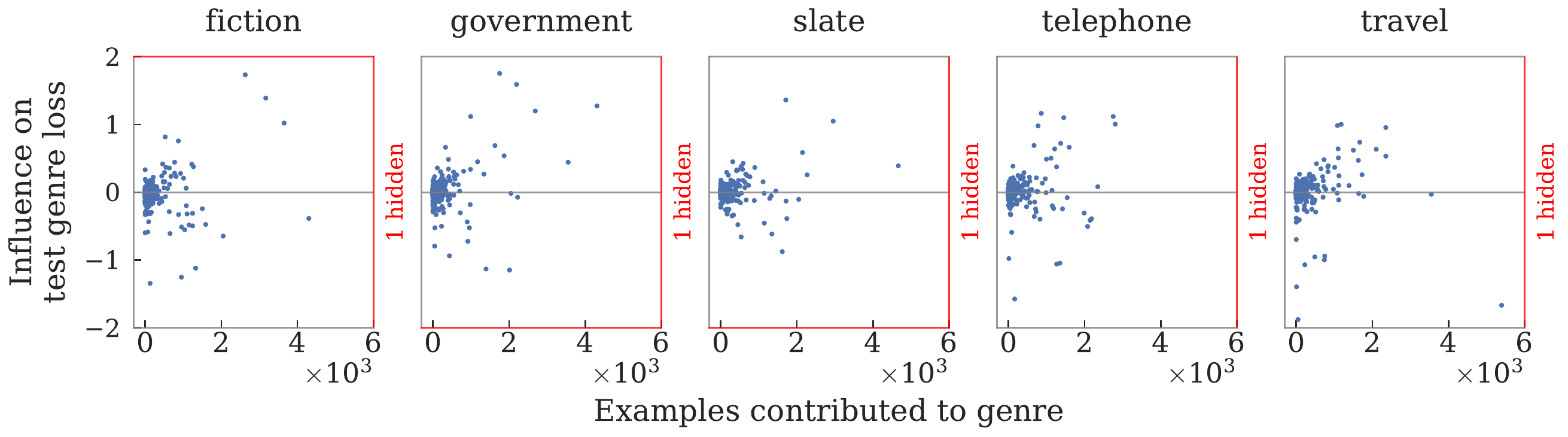}
  \caption{Influence on a genre does not correlate with number of contributions to that genre.
  }
  \label{fig:mnli-genre-contrib}
\end{figure}

The most prolific crowdworker contributed 35,000 examples and had large negative influence on the test set. A closer analysis revealed that they had positive influence on the fiction genre but lowered performance on many other genres, despite contributing roughly equally to each genre. This genre-specific trend tended to hold more broadly among the workers: there appear to be two categories of genres (fiction, facetoface, nineeleven vs. travel, government, verbatim, letters, oup) such that each worker tended to have positive influence on all genres in one category and negative influence on all genres in the other (\reffig{mnli-genre-categories}). Moreover, the number of examples a worker contributed to a given genre was not a good indicator for their influence on that genre (\reffig{mnli-genre-contrib}).

\section{Additional analysis on influence vs. actual effect on a test point}\label{sec:supp-theory}

\subsection{Counterexamples}\label{sec:theory-cex}

For \reffig{cex}, we constructed two binary datasets in which the influence of a
certain class of subsets on the test prediction of a single test point exhibits
pathological behavior.

\paragraph{Rotation effect.}
In \reffig{cex}-Left, our aim was to show that there can be a dataset with subsets such that the cone constraint discussed in \refsec{theory-test} does not hold.

The rotation effect described in \refcor{xtest} is due to the
angular difference between the change in parameters predicted by the influence approximation,
$\diw \eqdef \Ho^{-1} g_\bone(w) = \Ho^{-\frac 1 2} v_w$,
and the change in parameters predicted by the Newton approximation,
$\dnw = \Ho^{-\frac 1 2} \bigl(D(w) + I\bigr) v_w$.
If $\diw$ and $\dnw$ are linearly independent,
then for any pair of target values $a, b \in \R$,
we can find some $\xtest$ such that $\predI(w) = \xtest^\top \diw = a$ and $\nI(w  = \xtest^\top \dnw = b$.

To exploit this, we constructed the \emph{MoG} dataset as an equal mixture of two standard (identity covariance) Gaussian distributions in $\R^{60}$, one for each class, and with means $(-1/2, 0, \ldots, 0)$ and $(1/2, 0, \ldots, 0)$, respectively. In particular:
\begin{enumerate}
  \item We sampled $60$ examples from each class for a total of $n = 120$ training points, and set the regularization strength $\lambda = 0.001$.
  \item We then computed $\diw$ and $\dnw$ for each pair of training points and chose the $120$ pairs of training points with the largest angles between $\diw$ and $\dnw$.
  \item Finally, we solved a
  least-squares optimization problem to find $\xtest$ for which $\predI(w)$ and
  $\nI(w)$ are approximately decorrelated.
\end{enumerate}

Note that we adversarially chose which subsets to study in this counterexample, since our main goal was to show that there existed subsets for which the cone constraint did not hold.
For the next counterexample, we instead study all possible subsets in the restricted setting of removing copies of single points.

\paragraph{Scaling effect.} In \reffig{cex}-Right, our aim was to construct a dataset such that even if we only removed subsets comprising copies of single distinct points, a low influence need not translate into a low actual effect.

To do so, we constructed the \emph{Ortho} dataset that contains 2 repeated points of opposite classes on each of the 2 canonical axes of $\R^2$ (for a total of 4 distinct points). By varying their
relative distances from the origin, we can control the influence of removing one of these points as well as the rate that the scaling factor $d(w)$ from \refprop{single} grows as we remove more copies of the same point.
Furthermore, because the axes are orthogonal, we can control $d(w)$ independently for each repeated
point. We fix the test point $\xtest = (1, 1)$.
Maximizing $d(w)$ for one axis
and minimizing it for the other produces the two distinct lines in
\reffig{cex}-Right.

\subsection{Scaling effects when removing multiple points}\label{sec:theory-D}
In the general setting of removing subsets of different points, the analogous
failure case to a varying scaling factor $d(w)$ (\reffig{cex}-Right) is the
varying scaling effect that the error matrix $D(w)$ in \refprop{newton-inf} can have on
different subsets $w$. The range of this effect is bounded by the spectral norm
of $D(w)$. This norm is precisely equal to $d(w)$ in the single-point setting,
and it is large when we remove a subset $w$ whose Hessian
$H_\bone(w)$ is almost as large as the full Hessian $\Ho$ in some direction.
As with $d(w)$, the spectral norm of $D(w)$ decreases with $\lambda$ (\refprop{newton-inf}), so as regularization increases, we expect that the influence of a group will track its actual effect more accurately.

\subsection{The relationship between influence and actual effect on the loss of a test point}\label{sec:theory-loss}
In the margin-based setting, the loss $\ell(\xtest, \ytest; \theta)$ is a monotone function of the linear prediction $\theta^\top \xtest$. Thus, measuring $f(\theta) = \ell(\xtest, \ytest; \theta)$ will display the same rank correlation as measuring $f(\theta) = \theta^\top \xtest$ above,
so the same results about correlation in the test prediction setting carry over.

However, the second-order $f$-curvature term $\inv{2} \dnw^\top \nt^2 f(\hto) \dnw$ from \refprop{newton-inf} is always non-negative, even if the influence is negative. Under the assumption that $\Errf(w)$ and $\Errn(w)$ are both small
because they decay as $O(1/\lambda^3)$,
this implies that underestimation is only preserved when the influence is positive, as we observed empirically in \reffig{acc}-Mid.

\section{Proofs}\label{sec:proofs}
\subsection{Notation}\label{sec:notation}
We first review the notation given in \refsec{setting} and introduce new definitions that will be useful in the sequel. We define the empirical risk as
\begin{align*}
  L_s(\theta) \eqdef \left[ \sum_{i=1}^n s_i \ell(x_i, y_i; \theta)\right] + \frac{\lambda}{2} \|\theta\|_2^2,
\end{align*}
such that the optimal parameters are $\hat\theta(s) \eqdef {\arg\min}_{\theta\in\Theta} L_s(\theta)$.

Given sample weight vectors $r, s \in \R^n$, we define the derivatives
\begin{align*}
  g_r(s) &\eqdef \sum_{i=1}^n s_i\nabla_\theta \ell(x_i, y_i; \hat\theta(r))\\
  H_r(s) &\eqdef \sum_{i=1}^n s_i \nabla^2_\theta \ell(x_i, y_i; \hat\theta(r)).
\end{align*}
If the argument $s$ is omitted, it is assumed to be equal to $r$. For example,
\begin{align*}
  H_\bone \eqdef \sum_{i=1}^n \nabla^2_\theta \ell(x_i, y_i; \hto).
\end{align*}
If $H$ has a $\lambda$ subscript, then we add $\lambda I$. For example,
\begin{align*}
  \Ho \eqdef H_\bone + \lambda I.
\end{align*}

For a given dataset, we define the following constants:
\begin{align*}
  C_\ell &= \max_{1 \leq i \leq n} \twonorm{\nabla_\theta \ell(x_i,y_i,\hto)},\\
  \lmin &= \text{smallest singular value of}\ H_\bone,\\
  \lmax &= \text{largest singular value of}\ H_\bone.
\end{align*}

To avoid confusion with the vector 2-norm, we will use the operator norm $\opnorm{\cdot}$ to denote the matrix 2-norm.

In the sequel, we study the order-3 tensor $\nt^3 f(\hto)$. We define its product with a vector (which returns a matrix) as a contraction along the last dimension:
\begin{equation*}
  \left\langle \nt^3 f(\hto), v \right\rangle_{ij} \eqdef \sum_k \frac{\partial^3 f(\hto)}{\partial \theta_i \partial \theta_j \partial \theta_k} v_k.
\end{equation*}

\subsection{Assumptions}\label{sec:assumptions}
We make the following assumptions on the derivatives of the loss $\ell(x, y, \theta)$
and the evaluation function $f(\theta)$.
\begin{assumption}[Positive-definiteness and Lipschitz continuity of $H$]\label{assump:l-const}
  The loss $\ell(x, y, \theta)$ is convex and twice-differentiable in $\theta$, with positive regularization $\lambda > 0$.
  Further, there exists $C_H\in\R$ such that
  \[
    \opnorm{\nt^2 \ell(x, y, \theta_1) - \nt^2\ell(x, y, \theta_2)} \leq C_H \twonorm{\theta_1 - \theta_2}
  \]
  for all $(x, y) \in \sX \times \sY$ and $\theta_1, \theta_2 \in \Theta$. This is a bound on the third derivative of $\ell$, if it exists.
\end{assumption}
\begin{assumption}[Bounded derivatives of $f$]\label{assump:q-const}
  $f(\theta)$ is thrice-differentiable, with $C_f, C_{f, 3} \in \R$ such that
  \begin{align*}
    C_{f} &= \sup_{\theta \in \Theta} \twonorm{\nt f(\theta)},
    &C_{f,3} &= \sup_{v \in \Theta, \twonorm{v} = 1} \opnorm{
    \left\langle \nt^3 f(\hto), v \right\rangle}.
  \end{align*}
\end{assumption}
These assumptions apply to all the results that follow below.

\subsection{Bounding the error of the one-step Newton approximation}\label{sec:proof-newton}

\begin{repproposition}{prop:newton-actual}[Restated]
  Let the Newton error be $\Errn(w) \eqdef \actI(w) - \nI(w)$.
  Then under \refassumps{l-const}{q-const},
  \begin{align*}
      \abs{\Errn(w)} \leq \frac{n \|w\|_1^2 C_f C_H C_\ell^2}{(\lmin + \lambda)^3}.
  \end{align*}
  $\Errn(w)$ only involves third-order or higher derivatives of the loss, so it is 0 for quadratic losses.
\end{repproposition}
\begin{proof}
  This proof is adapted to our setting from the standard analysis of the Newton method in convex optimization \citep{boyd2004convex}.

  First, note that $\Errn(w) = \actI(w) - \nI(w) = f(\htow) - f(\htn(\ow))$.
  We will bound the norm of the difference of the parameters $\twonorm{\htow - \htn(\ow)}$; the desired bound on $f$ then follows from the assumption that $f$ has gradients bounded by $C_f$ and is therefore Lipschitz.

  Since $L_\ow(\theta)$ is strongly convex (with parameter $\lmin + \lambda$) and minimized by $\htow$, we can bound the distance $\twonorm{\htow - \htn(\ow)}$ in terms of the norm of the gradient at $\htn(\ow)$:
  \begin{align*}
    \twonorm{\htow - \htn(\ow)} &\leq \frac{2}{\lmin + \lambda}\twonorm{\nt L_\ow\left(\htn(\ow)\right)}.
  \end{align*}

  Therefore, the problem reduces to bounding $\twonorm{\nt L_\ow\left(\htn(\ow)\right)}$.

  We start by expressing the Newton step $\dnw$ in terms of the first and second derivatives of the empirical risk $L_\ow(\theta)$:
  \begin{align*}
    g_\bone(w) &= \sum_{i=1}^n w_i \nt\ell(x_i, y_i; \hto)\\
    &= -\sum_{i=1}^n (1 - w_i) \nt\ell(x_i, y_i; \hto)\\
    &= -\nt L_\ow(\hto),\\
    \Ho(\ow) &= \sum_{i=1}^n (1 - w_i) \nt^2 \ell(x_i, y_i; \hto)\\
    &= \nt^2 L_\ow(\hto),\\
    \dnw &= \Ho(\ow)^{-1} g_\bone(w)\\
    &= -\left[ \nt^2 L_\ow(\hto) \right]^{-1} \nt L_\ow(\hto),
  \end{align*}
  where the second equality for $g_\bone(w)$ comes from the fact that at the optimum $\hto$, the sum of the gradients $\sum_{i=1}^n \nt\ell(x_i, y_i; \hto)$ is 0.

  With these expressions, we bound the norm of the gradient $\nt L_\ow(\htn(\ow))$:
  \begin{align*}
    &\twonorm{\nt L_\ow\left(\htn(\ow)\right)}\\
    &= \twonorm{\nt L_\ow\left(\hto + \dnw\right)}\\
    &= \twonorm{\nt L_\ow\left(\hto + \dnw\right)
       - \nt L_\ow\left(\hto\right)
       - \nt^2 L_\ow\left(\hto\right) \dnw}\\
    &= \twonorm{\int_0^1 \left(\nt^2 L_\ow\left(\hto + t\dnw \right) - \nt^2 L_\ow\left(\hto\right)\right) \dnw \ dt}\\
    &\leq \frac{n C_H}{2} \twonorm{\dnw}^2\\
    &= \frac{n C_H}{2} \twonorm{\left[ \nt^2 L_\ow(\hto) \right]^{-1} \nt L_\ow(\hto)}^2\\
    &\leq \frac{n C_H}{2(\lmin + \lambda)^2} \twonorm{\nt L_\ow(\hto)}^2\\
    &\leq \frac{n \onenorm{w}^2 C_H C_\ell^2}{2(\lmin + \lambda)^2}.
  \end{align*}
  Putting together the successive bounds gives the result.
\end{proof}

\subsection{Characterizing the difference between the Newton approximation and influence}\label{sec:proof-decomp}
Before proving \refprop{newton-inf}, we first prove a lemma about the spectrum of the error matrix $D(w)$.
\begin{lemma}\label{lem:sv-bound-d}
The matrix $D(w) \eqdef \bigl(I - \Ho^{-\inv{2}}H_\bone(w)\Ho^{-\inv{2}}\bigr)^{-1} - I$ has singular values bounded between 0 and $\frac{\lmax}{\lambda}$.
\end{lemma}
\begin{proof}
We first show that $\Ho^{-\inv{2}}H_\bone(w)\Ho^{-\inv{2}}$ has singular values bounded between 0 and $\frac{\lmax}{\lmax + \lambda}$.
The lower bound of 0 comes from the fact that $\Ho^{-\inv{2}} H_\bone(w) \Ho^{-\inv{2}}$ is symmetric and $H_\bone(w) \succeq 0$.

To show the upper bound, first note that $H_\bone(w) \preceq H_\bone(w) + H_\bone(\bone - w) = H_\bone$ (recalling that $w\in\{0,1\}^n$),
and let $U \Sigma U^\top$ be the singular value decomposition of $H_\bone$. Since $\Ho = H_\bone + \lambda I$, we have
\begin{align*}
  \Ho^{-\inv{2}} H_\bone(w) \Ho^{-\inv{2}} &=
  \bigl(H_\bone + \lambda I\bigr)^{-\inv{2}} H_\bone(w) \bigl(H_\bone + \lambda I\bigr)^{-\inv{2}}\\
  &\preceq \bigl(H_\bone + \lambda I\bigr)^{-\inv{2}} H_\bone \bigl(H_\bone + \lambda I\bigr)^{-\inv{2}}\\
  &= \bigl(U (\Sigma + \lambda I) U^\top\bigr)^{-\inv{2}} U \Sigma U^\top \bigl(U (\Sigma + \lambda I) U^\top\bigr)^{-\inv{2}}\\
  &= U (\Sigma + \lambda I)^{-\inv{2}} \Sigma (\Sigma + \lambda I)^{-\inv{2}} U^\top,
\end{align*}
so its maximum singular value is upper bounded by $\frac{\lmax}{\lmax + \lambda}$.

The bound on the singular values of $\Ho^{-\inv{2}}H_\bone(w)\Ho^{-\inv{2}}$ implies that the singular values of $I - \Ho^{-\inv{2}}H_\bone(w)\Ho^{-\inv{2}}$ lie in $\left[\frac{\lambda}{\lmax + \lambda}, 1\right]$.
In turn, this implies that the singular values of $\bigl(I - \Ho^{-\inv{2}}H_\bone(w)\Ho^{-\inv{2}}\bigr)^{-1}$ lie in $\left[1, \frac{\lmax + \lambda}{\lambda}\right]$. Subtracting 1 from each end (for the identity matrix) gives the desired result.
\end{proof}

\begin{repproposition}{prop:newton-inf}[Restated]
  Under \refassumps{l-const}{q-const}, the Newton-influence error $\Errni(w)$ is
  \begin{align*}
    \Errni(w) &= \nt f(\hto)^\top \Ho^{-\inv{2}} D(w) \Ho^{-\inv{2}} g_\bone(w)
    \ + \ \underbrace{\inv{2} \dnw^\top \nt^2 f(\hto) \dnw + \Errf(w),}_\text{
      Error from curvature of $f(\cdot)$
    }
  \end{align*}
  with $D(w) \eqdef \bigl(I - \Ho^{-\inv{2}}H_\bone(w)\Ho^{-\inv{2}}\bigr)^{-1} - I$
  and $H_\bone(w) \eqdef \sum_{i=1}^n w_i \nt^2\ell(x_i, y_i; \hto)$.
  The \emph{error matrix} $D(w)$ has eigenvalues between 0 and $\frac{\lmax}{\lambda}$,
  where $\lmax$ is the largest eigenvalue of $H_\bone$.
  The residual term $\Errf(w)$ captures the error due to third-order derivatives of $f(\cdot)$ and is bounded by $\abs{\Errf(w)} \leq {\|w\|_1^3 C_{f, 3}C_\ell^3}/{6(\lmin + \lambda)^3}$.
\end{repproposition}
\begin{proof}
  From the second-order Taylor expansion of $f$ about $\hto$, there exists $0 \leq \xi \leq 1$ such that
  \begin{align*}
    \nI(w) &= f(\htn(\ow)) - f(\hto)\\
    &= f(\hto + \dnw) - f(\hto)\\
    &= \nt f(\hto)^\top \dnw + \inv{2} \dnw^\top \nt^2 f(\hto) \dnw +\\
    &\quad \inv{6} \dnw^\top \left\langle \nt^3 f(\hto + \xi\dnw), \dnw \right\rangle \dnw\\
    &= \nt f(\hto)^\top \dnw + \inv{2} \dnw^\top \nt^2 f(\hto) \dnw + \Errf(w),
    \numberthis \label{eqn:newton-inf-proof-1}
  \end{align*}
  where we define $\Errf(w) \eqdef \inv{6} \dnw^\top \left\langle \nt^3 f(\hto + \xi\dnw), \dnw \right\rangle \dnw$ to be the error due to third-order and higher derivatives of $f$.

  We can express the difference between the first-order Taylor term $\nt f(\hto)^\top \dnw$ above and the first-order influence approximation $\predI(w) = q'_w(0) = \nt f\bigl(\hto\bigr)^\top \Ho^{-1} g_\bone(w)$ as
  \begin{align*}
    &\nt f(\hto)^\top \dnw - \predI(w)\\
    &=\nt f(\hto)^\top \dnw - \nt f\bigl(\hto\bigr)^\top \Ho^{-1} g_\bone(w)\\
    &=\nt f(\hto)^\top \left(\Ho(\ow)^{-1} - \Ho^{-1}\right) g_\bone(w)\\
    &=\nt f(\hto)^\top \left( \big(\Ho - H_\bone(w)\big)^{-1} - \Ho^{-1}\right) g_\bone(w)\\
    &=\nt f(\hto)^\top \Ho^{-\inv{2}} \left(\Ho^{\inv{2}}\big(\Ho - H_\bone(w)\big)^{-1}\Ho^{\inv{2}} - I\right) \Ho^{-\inv{2}} g_\bone(w)\\
    &=\nt f(\hto)^\top \Ho^{-\inv{2}} \left(\big(I - \Ho^{-\inv{2}}H_\bone(w)\Ho^{-\inv{2}}\big)^{-1} - I\right) \Ho^{-\inv{2}} g_\bone(w)\\
    &=\nt f(\hto)^\top \Ho^{-\inv{2}} D(w) \Ho^{-\inv{2}} g_\bone(w)
    \numberthis \label{eqn:newton-inf-proof-2}.
  \end{align*}
  Substituting \refeqn{newton-inf-proof-2} into \refeqn{newton-inf-proof-1},
  we have that
  \begin{align*}
    \nI(w) - \predI(w)
    &= \nt f(\hto)^\top \Ho^{-\inv{2}} D(w) \Ho^{-\inv{2}} g_\bone(w)\\
    &\quad + \inv{2} \dnw^\top \nt^2 f(\hto) \dnw
    + \Errf(w),
  \end{align*}
  as desired.

  We can bound $\Errf(w)$ as follows:
  \begin{align*}
    \abs{\Errf(w)} &\leq
    \frac{C_{f, 3}}{6}\twonorm{\dnw}^3\\
    &\leq \frac{\onenorm{w}^3 C_{f, 3}C_\ell^3}{6(\lmin + \lambda)^3}.
  \end{align*}
  Applying \reflem{sv-bound-d} to bound the spectrum of $D(w)$ completes the proof.
\end{proof}

\subsection{The influence on self-loss}\label{sec:proof-selfloss}
We first state two linear algebra facts that will be useful in the sequel.
\begin{lemma}\label{lem:sv-bound}
  Let $A \succ 0, B \succeq 0 \in \R^{d \times d}$ be a pair of symmetric positive-definite and positive-semidefinite matrices, respectively. Let $\sigma_{A, 1}$ be the largest eigenvalue of $A$, $\sigma_{A, d}$ the smallest eigenvalue of $A$, and similarly let $\sigma_{B, 1}$ and $\sigma_{B, d}$ be the largest and smallest eigenvalues of $B$, respectively. Then
  \begin{align*}
    \frac{\sigma_{B, d}}{\sigma_{A, 1}} I
    \preceq A^{-\inv{2}} B A^{-\inv{2}}
    \preceq \frac{\sigma_{B, 1}}{\sigma_{A, d}} I.
  \end{align*}
\end{lemma}
\begin{proof}
  Note that $\inv{\sigma_{A, 1}}$ is the smallest eigenvalue of $A^{-1}$, while $\inv{\sigma_{A, d}}$ is its largest. The lemma follows from the fact that the smallest singular value of the product of two matrices is lower bounded by the product of the smallest singular values of each matrix, and similarly the largest singular value of the product is upper bounded by the product of the largest singular values of each matrix.
\end{proof}

The next fact is a consequence of the variational definition of eigenvalues.
\begin{lemma}\label{lem:quadform-bound}
  Given a symmetric matrix $A \in \R^{d \times d}$ and a vector $v \in \R^d$, we have the following bounds on the quadratic form $v^\top A v$:
  \[
  \sigma_d \twonorm{v}^2 \leq v^\top A v \leq \sigma_1 \twonorm{v}^2,
  \]
  where $\sigma_d$ is the smallest eigenvalue of $A$, and $\sigma_1$ is the largest.
\end{lemma}

We are now ready to analyze the effect of removing a subset $w$ of $k$ training points on the total loss on those $k$ points.
\begin{repproposition}{prop:selfloss}[Restated]
  Under \refassumps{l-const}{q-const}, the influence on the self-loss $f(\theta) = \sum_{i=1}^n w_i \ell(x_i, y_i; \theta)$ obeys
  \begin{align*}
    \predI(w) + \Errf(w) &\leq \nI(w)
    \leq \left(1 + \frac{3\lmax}{2\lambda} + \frac{\lmax^2}{2\lambda^2}\right) \predI(w) + \Errf(w).
  \end{align*}
\end{repproposition}
\begin{proof}
  Since $f(\theta) = \sum_{i=1}^n w_i \ell(x_i, y_i; \theta)$, we have that
  \begin{align*}
    \nabla_\theta f(\hto) &= \sum_{i=1}^n w_i \nabla_\theta \ell(x_i, y_i; \hto)\\
    &= g_\bone(w),\\
    \nabla_\theta^2 f(\hto) &= \sum_{i=1}^n w_i \nabla_\theta^2 \ell(x_i, y_i; \hto)\\
    &= H_\bone(w).
  \end{align*}
  Substituting these and $\dnw = \Ho(\ow)^{-1} g_\bone(w)$ into \refprop{newton-inf}, we obtain
  \begin{align*}
    &\nI(w) - \predI(w) - \Errf(w)\\
    &= \nt f(\hto)^\top \Ho^{-\inv{2}} D(w) \Ho^{-\inv{2}} g_\bone(w)
      + \inv{2} \dnw^\top \nt^2 f(\hto) \dnw\\
    &= g_\bone(w)^\top \Ho^{-\inv{2}} D(w) \Ho^{-\inv{2}} g_\bone(w)
      + \inv{2} g_\bone(w)^\top \Ho(\ow)^{-1} H_\bone(w) \Ho(\ow)^{-1} g_\bone(w)\\
      &= g_\bone(w)^\top \Ho^{-\inv{2}} \left[ D(w) + \underbrace{\inv{2}\Ho^{\inv{2}}\Ho(\ow)^{-1} H_\bone(w) \Ho(\ow)^{-1} \Ho^{\inv{2}}}_{\eqdef \Lambda(w)}\right] \Ho^{-\inv{2}} g_\bone(w),
  \end{align*}
  where $D(w) \eqdef \bigl(I - \Ho^{-\inv{2}}H_\bone(w)\Ho^{-\inv{2}}\bigr)^{-1} - I$ has singular values bounded between 0 and $\frac{\lmax}{\lambda}$.
  From \reflem{sv-bound}, $\Lambda(w)$ has singular values bounded between $0$ and
  $\frac{\lmax(\lmax + \lambda)}{2\lambda^2}$.

  Applying \reflem{quadform-bound} and using $\predI(w) = g_\bone(w)^\top \Ho^{-1} g_\bone(w)$, we obtain
  \begin{align*}
    0 &\leq
    g_\bone(w)^\top \Ho^{-\inv{2}} \left[ D(w) + \Lambda(w) \right] \Ho^{-\inv{2}} g_\bone(w)\\
    &\leq \left(\frac{\lmax}{\lambda} + \frac{\lmax(\lmax + \lambda)}{2\lambda^2}\right) g_\bone(w)^\top \Ho^{-1} g_\bone(w)\\
    &= \left(\frac{3\lmax}{2\lambda} + \frac{\lmax^2}{2\lambda^2}\right) \predI(w),
  \end{align*}
  which gives us
  \begin{align*}
      \predI(w) + \Errf(w) \leq \nI(w) \leq \left(1 + \frac{3\lmax}{2\lambda} + \frac{\lmax^2}{2\lambda^2}\right) \predI(w) + \Errf(w).
  \end{align*}
  Note that $\tErrf(w) \eqdef \frac{\lmax^2}{2\lambda^2} \predI(w)$ can be bounded as
  \begin{align*}
    \abs{\tErrf(w)} &= \abs{\frac{\lmax^2}{2\lambda^2} \predI(w)}\\
    &\leq \frac{\lmax^2}{2\lambda^2} \cdot \abs{\predI(w)}\\
    &\leq \frac{\lmax^2}{2\lambda^2} \cdot \abs{g_\bone(w)^\top \Ho^{-1} g_\bone(w)}\\
    &\leq \frac{\onenorm{w}^2 C_\ell^2 \lmax^2}{2 (\lmin + \lambda) \lambda^2},
  \end{align*}
  and $\Errn(w) = \actI(w) - \nI(w)$ grows as $O(1/\lambda^3)$ from \refprop{newton-actual},
  so we can also write
  \begin{align*}
      \predI(w) + O\Bigl(\inv{\lambda^3}\Bigr) \leq \actI(w) \leq \left(1 + \frac{3\lmax}{2\lambda}\right) \predI(w) + O\Bigl(\inv{\lambda^3}\Bigr).
  \end{align*}

\end{proof}

\subsection{The influence on a test point}
\begin{repcorollary}{cor:xtest}[Restated]
  Suppose $f(\theta) = \theta^\top \xtest$,
  and define $\vtest \eqdef \Ho^{-\inv{2}} \xtest$ and $v_w \eqdef \Ho^{-\inv{2}} g_\bone(w)$.
  Then under \refassumps{l-const}{q-const}, $\nI(w) = \predI(w) + \vtest^\top D(w) v_w$,
  where $D(w) = \bigl(I - \Ho^{-\inv{2}}H_\bone(w)\Ho^{-\inv{2}}\bigr)^{-1} - I$ is the \emph{error matrix} from \refprop{newton-inf}.
\end{repcorollary}
\begin{proof}
  Since $f(\theta) = \theta^\top \xtest$ is linear, we have that for any $\theta \in \Theta$,
  \begin{align*}
    \nabla_\theta f(\theta) &= \xtest,\\
    \nabla_\theta^2 f(\theta) &= 0,\\
    C_{f, 3} &= 0.
  \end{align*}
  This in turn implies that $\Errf(w) = 0$. Substituting these expressions into  \refprop{newton-inf} gives us the desired result.
\end{proof}

\begin{repproposition}{prop:single}[Restated]
  Consider a binary classification setting with $y \in \{-1, +1\}$ and a margin-based model with loss
  $\ell(x, y; \theta) = \phi(y\theta^\top x)$ for some $\phi:\R\to\R_+$.
  Suppose $f(\theta) = \theta^\top \xtest$
  and that the subset $w$ comprises $\|w\|_1$ identical copies of the training point $(x_w, y_w)$.
  Then under \refassumps{l-const}{q-const},
  the Newton approximation $\nI(w)$ is related to the influence $\predI(w)$ according to
  \begin{equation*}
    \nI(w) = \frac{\predI(w)}
        {1 - \onenorm{w} \cdot \phi''(y_w \hto^\top x_w) \cdot x_w^\top \Ho^{-1} x_w}.
  \end{equation*}
  This implies the Newton approximation $\nI(w)$ is bounded between $\predI(w)$ and
  $\bigl(1 + \frac{\lmax}{\lambda}\bigr) \predI(w)$.
\end{repproposition}
\begin{proof}
  From \refcor{xtest},
  \begin{align*}
    \nI(w) &= \predI(w) + \xtest^\top \Ho^{-\inv{2}} D(w) \Ho^{-\inv{2}} g_\bone(w).
  \end{align*}
  With the additional assumptions on $w$ and $\ell(x, y; \theta)$, we have that
  \begin{align*}
    \nabla_\theta \ell(x, y; \theta) &= y \phi'(y\theta^\top x) x,\\
    g_\bone(w) &= \sum_{i=1}^n w_i \nabla_\theta \ell(x_i, y_i; \hto)\\
    &= \sum_{i=1}^n w_i y_i \phi'(y_i\hto^\top x_i) x_i\\
    &= \sum_{i=1}^n w_i y_i \phi_i' x_i\\
    &= \onenorm{w} y_w \phi_k' x_w,
  \end{align*}
  where in the last equality we use the assumption that we are removing $\onenorm{w}$ copies of the point $(x_w, y_w)$. Similarly,
  \begin{align*}
    \nabla_\theta^2 \ell(x, y; \theta) &= \phi''(y\theta^\top x) xx^\top,\\
    H_\bone(w) &= \sum_{i=1}^n w_i \nabla^2_\theta \ell(x_i, y_i; \hto)\\
    &= \sum_{i=1}^n w_i \phi''(y_i\hto^\top x_i) x_i x_i^\top\\
    &= \onenorm{w} \phi''_k x_w x_w^T.
  \end{align*}
  We thus have
  \begin{align*}
    D(w) &= \bigl(I - \Ho^{-\inv{2}}H_\bone(w)\Ho^{-\inv{2}}\bigr)^{-1} - I\\
    &= \Bigl(I - \Ho^{-\inv{2}}
    \onenorm{w} \phi''_k x_w x_w^T
    \Ho^{-\inv{2}}\Bigr)^{-1} - I\\
    &= \frac{
      \Ho^{-\inv{2}}
      \onenorm{w} \phi''_k x_w x_w^T
      \Ho^{-\inv{2}}}
    {1 - \onenorm{w} \phi''_k x_w^T \Ho^{-1} x_w}\\
    &= \frac{
    \onenorm{w} \phi''_k \Ho^{-\inv{2}}
       x_w x_w^T
      \Ho^{-\inv{2}}}
    {1 - \onenorm{w} \phi''_k x_w^T \Ho^{-1} x_w},
  \end{align*}
  where the third equality comes from the Sherman-Morrison formula.
  Substituting $D(w)$ into \refcor{xtest}, we obtain
  \begin{align*}
    \nI(w) &= \predI(w) + \xtest^\top \Ho^{-\inv{2}} D(w) \Ho^{-\inv{2}} g_\bone(w)\\
    &= \predI(w) +
      \frac{
        \onenorm{w} \phi''_k \xtest^\top \Ho^{-1}
        x_w x_w^T
        \Ho^{-1} g_\bone(w)
        }
        {1 - \onenorm{w} \phi''_k x_w^T \Ho^{-1} x_w}\\
    &= \predI(w) +
      \frac{
        \xtest^\top \Ho^{-1} \onenorm{w} y_w \phi_k' x_w
        \cdot \onenorm{w} \phi''_k x_w^T \Ho^{-1} x_w
        }
        {1 - \onenorm{w} \phi''_k x_w^T \Ho^{-1} x_w}\\
      &= \predI(w) +
        \frac{
          \xtest^\top \Ho^{-1} g_\bone(w)
          \cdot \onenorm{w} z_k^T \Ho^{-1} z_k
          }
          {1 - \onenorm{w} \phi''_k x_w^T \Ho^{-1} x_w}\\
        &= \predI(w) +
          \frac{
            \predI(w)
            \cdot \onenorm{w} z_k^T \Ho^{-1} z_k
            }
            {1 - \onenorm{w} \phi''_k x_w^T \Ho^{-1} x_w}\\
          &= \frac{\predI(w)}
              {1 - \onenorm{w} \phi''_k x_w^T \Ho^{-1} x_w}.
  \end{align*}
  To bound the denominator, we first use the trace trick to rearrange terms
  \begin{align*}
    \onenorm{w} \phi''_k x_w^T \Ho^{-1} x_w &=
    \tr\left(\onenorm{w} \phi''_k x_w^T \Ho^{-1} x_w\right)\\
    &= \tr\left( \Ho^{-\inv{2}} \onenorm{w} \phi''_k x_w x_w^T \Ho^{-\inv{2}}\right)\\
    &= \tr\left(\Ho^{-\inv{2}}H_\bone(w)\Ho^{-\inv{2}} \right).
  \end{align*}
  Since $\Ho^{-\inv{2}}H_\bone(w)\Ho^{-\inv{2}}$ has rank one under our assumptions, it only has at most one non-zero eigenvalue. We can therefore apply \reflem{sv-bound-d} to conclude that
  \begin{align*}
    \onenorm{w} \phi''_k x_w^T \Ho^{-1} x_w &= \tr\left(\Ho^{-\inv{2}}H_\bone(w)\Ho^{-\inv{2}} \right)\\
    &\leq \frac{\lmax}{\lmax + \lambda},
  \end{align*}
  which in turn implies that
    $1 - \onenorm{w} \phi''_k x_w^T \Ho^{-1} x_w
    \geq \frac{\lambda}{\lmax + \lambda}$,
  so
  \begin{align*}
    \inv{1 - \onenorm{w} \phi''_k x_w^T \Ho^{-1} x_w}
    \leq \frac{\lmax + \lambda}{\lambda}=1 + \frac{\lmax}{\lambda}.
  \end{align*}

\end{proof}

\end{document}